\documentclass[preprint]{elsarticle}

\usepackage{amssymb}
\usepackage{amsthm}
\usepackage{amsmath}
\usepackage{epsfig}
\usepackage{grffile}
\usepackage{graphicx}
\usepackage[caption=false]{subfig}

\usepackage{color}
\newcommand{\xxx}[1]{ \textcolor{black}{#1} }

\usepackage{algorithm2e}
\usepackage{url}
\usepackage{array}

\usepackage{multirow}
\usepackage{hhline}

\DeclareMathOperator*{\argmax}{\arg\!\max}

\journal{Medical Image Analysis}

\begin{document}

\begin{frontmatter}



\title{Optimal Multi-Object Segmentation with Novel Gradient Vector Flow Based Shape Priors}


\author[uiowa]{Junjie Bai}
\author[uiowa]{Abhay Shah}
\author[uiowa]{Xiaodong Wu}

\address[uiowa]{Department of Electrical and Computer Engineering, The University of Iowa, Iowa City, IA, 52242, USA}

\begin{abstract}
Shape priors have been widely utilized in medical image segmentation to improve segmentation accuracy and robustness. A major way to encode such a prior shape model is to use a mesh representation, which 
%
%
is prone to causing self-intersection or mesh folding. Those problems require complex and expensive algorithms to mitigate. In this paper, we propose a novel shape prior directly embedded in the voxel grid space, based on gradient vector flows of a pre-segmentation. The flexible and powerful prior shape representation is ready to be  extended to simultaneously segmenting multiple interacting objects with minimum separation distance constraint. The problem is formulated as a Markov random field problem whose exact solution can be efficiently computed with a single minimum $s$-$t$ cut in an appropriately constructed graph. 

The proposed algorithm is validated on two multi-object segmentation applications: the brain tissue segmentation in MRI images, and the bladder/prostate segmentation in CT images. Both sets of experiments show superior or competitive performance of the proposed method to other state-of-the-art methods. 
\end{abstract}


\end{frontmatter}


\section{Introduction}

Shape priors have been widely used in medical image segmentation due to the similar intensity profiles and weak boundaries between the target objects and background.
For example, the popular deformable model~\cite{Jalba2004,Weese2001,Martin2010,McInerney1996} and the LOGISMOS (layered optimal graph image segmentation of multiple objects and surfaces)~\cite{Wu2002,Oguz2014,Yin2010,Song2010miccai,Song2009} methods  use a mesh in the physical space to ensure that the resulting segmentation is roughly aligned to the initial model or the pre-segmentation. Recently, the star-shaped prior has been successfully encoded in the voxel grid space, which can be incorporated into the graph-based image segmentation methods with efficient global optimizers~\cite{gulshan_geodesic_2010,Veksler2008,Bai2014}.


\subsection{Shape Priors with a Mesh-Based Representation}
\label{subsec:shapePriorByMesh}
Deformable model first initializes a mesh representing the target object close to the desired location and orientation~\cite{Dornheim2006,Jalba2004,Weese2001}. The boundary and regional information are then used to evolve the mesh vertices along  their normal directions to adhere to the image content. An elastic force is also often used to limit the amount of warp to enforce the shape prior~\cite{Dornheim2006,Weese2001}. Such evolution is repeated multiple times until convergence. 

LOGISMOS has been widely used for multiple surface segmentation with the capability of enforcing mutual surface interactions, while still achieving globally optimal solutions~\cite{Wu2002,Li2006a,Yin2010,Song2009,Song2010miccai}. The method first builds a mesh based on an initial pre-segmentation or a shape model of the target object. The image is then resampled usually along the normal direction at each mesh vertex to form a graph node column. The graph nodes on each column represent the set of all possible target surface locations sampled at the normal direction at each mesh vertex. Various edges are then added among the nodes to ensure the smoothness of the surfaces, and the mutually interactive relationships between them.
The globally optimal surface locations are then computed by solving a single minimum $s$-$t$ cut in the constructed graph. 

\begin{figure}
	\centering
	\subfloat[column intersection]{\includegraphics[width=0.33\linewidth]{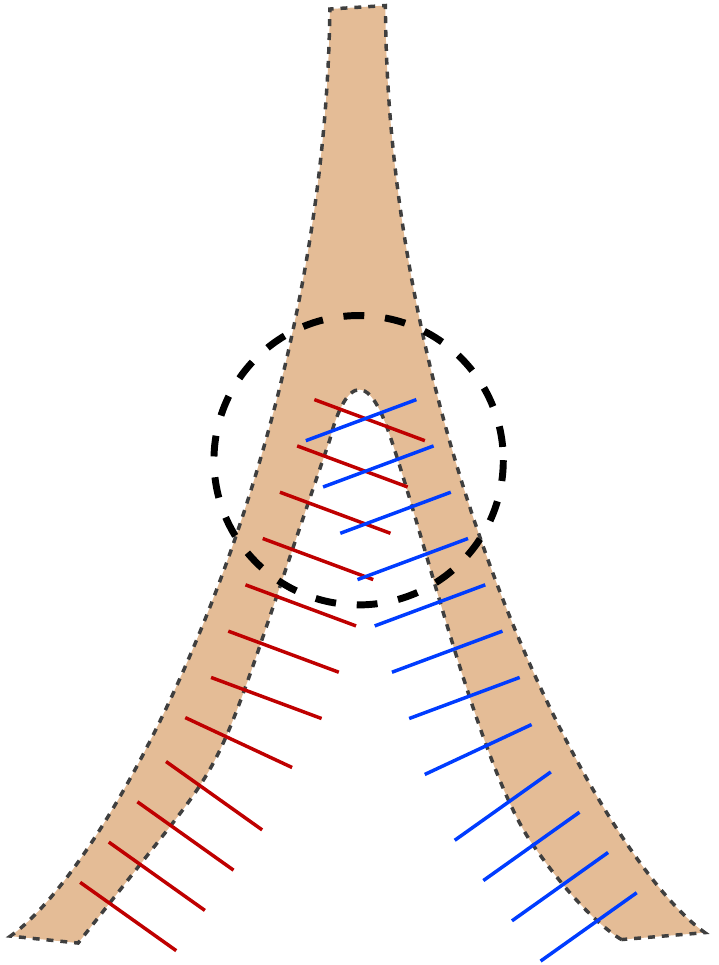}%
		\label{subfig:meshColumn}}
	\hspace{15pt}
	\subfloat[folded mesh]{\includegraphics[width=0.33\linewidth]{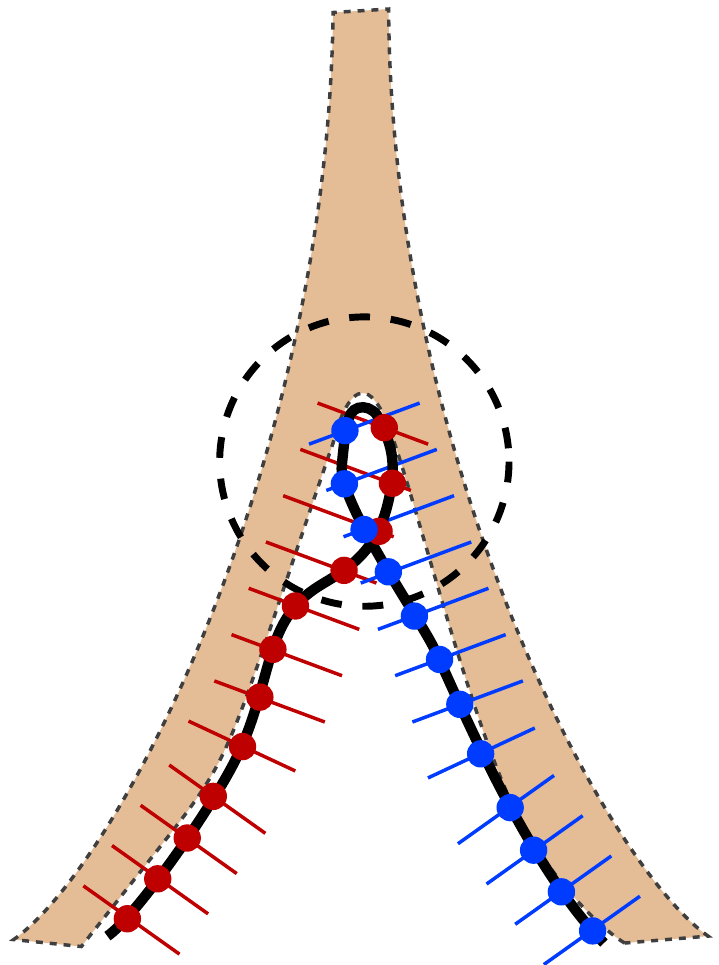}%
		\label{subfig:meshFolding}}
	\caption{Mesh folding at the bifurcation locations. \protect\subref{subfig:meshColumn} shows part of LOGISMOS graph columns built based on a pre-segmentation (light brown region). The red columns and blue columns are intersecting in the circled area. This may lead to folded mesh as shown in \protect\subref{subfig:meshFolding} in the solution. Although deformed model do not explicitly build such columns, the mesh folding may still happen during mesh evolving.}
	\label{fig:meshFolding}
\end{figure}

Although widely used in medical image segmentation, the mesh-based shape prior representation has several shortcomings in practice. Firstly, the mesh may be ``self-intersected'' or ``folded'' in the output solution, causing undesired topological changes from the prior shape. The problem is especially noticeable at the high curvature locations such as concavities and bifurcations (see Figure.\ref{fig:meshFolding}). To avoid this problem, the deformable method need to run expensive self-intersection detection and re-meshing algorithms in each evolution step~\cite{Park2001,Pons2007,Wang2008}. For LOGISMOS, various methods have been proposed to prevent the interference between graph node columns. The electrical field~\cite{Yin2009}, flow lines~\cite{Petersen2014} and gradient vector flow~\cite{Oguz2014} are used to build curved graph columns instead of the traditional straight-line columns along the normal directions. 
\xxx{Veni \textit{et al.}~\cite{Veni2013,Veni2013a} devised a flexible 3D lattice particle system to compute non-intersecting graph column trajectories. The graph columns for the same layer are generated with repulsive spring-like forces, while the columns for the interactions between different layers are computed with attractive spring-like forces. The whole spring potential system, if well balanced, could avoid mesh folding. However, solving the spring system is nontrivial. A local gradient descent method with asynchronous updates was used by the authors.}
The omnidirectional displacement method~\cite{Kainmueller2013} has also been proposed to distribute graph nodes uniformly in a sphere centered at each mesh vertex. However, the resulting MRF problem can no longer be solvable by the minimum $s$-$t$ cut algorithm, and the high computational complexity restricts its wide application. 

Secondly, it is difficult to enforce the interactions between the shape priors for  multiple objects. 
For deformable model method, an expensive mesh-collision detection algorithm has to be run at the end of each evolving iteration to prevent two close objects from intersecting each other~\cite{Klinder2009,Teschner2005}. For LOGISMOS-based methods, nesting surfaces must share the same base mesh in order to enforce the surface interaction constraints~\cite{Oguz2014}.
\xxx{For example, the same base mesh is used for segmenting both inner and outer walls of left atrium with soft smoothing penalties~\cite{Veni2013,Veni2013a}. When multiple meshes for different shape priors are used, a workaround must be proposed to register multiple meshes into the same physical space. In~\cite{Song2010miccai,Song2009}, Song \textit{et al.} proposed a heuristic method to merge the graph columns from the mesh for the prostate shape prior and the mesh for the bladder shape prior into a common ``interactive region'' in order to enforce the exclusion relationship of the two objects. To make use of multiple \emph{different} shape priors for interactive objects efficiently, we should embed the shape priors directly in the voxel space instead of separate mesh spaces. }


\subsection{Shape Priors Defined Directly in the Voxel Grid Space}
Another group of methods embed the shape prior directly in the voxel grid space, fundamentally eliminating the drawbacks of the mesh representation. Veksler \textit{et al.}~\cite{Veksler2008} introduced a flexible star-shaped prior which requires every point in the target object to be visible to a predefined \emph{star center} through a straight line totally within the object. The prior is enforced by adding into the graph cut method~\cite{Boykov2006} a set of edges pointing towards the star center; each edge is set a weight of $+\infty$. Those edge mimic the Euclidean rays originating from the star center. However, the discretized rays forming with those edges computed by heuristics may not behave like their Euclidean counterparts.
Bai \textit{et al.}~\cite{Bai2014} utilized the well-behaved \emph{consistent digital rays}~\cite{christ_consistent_2012,chun_consistent_2009} to systematically build a set of edges so that the formed discretized rays have a tight Hausdorff distance bound to their Euclidean counterparts. Gulshan {\em et al.}~\cite{gulshan_geodesic_2010} extended Veksler's star-shaped prior by using geodesic paths instead of Euclidean rays to define a star shaped object. \xxx{The geodesic paths are defined by combining the Euclidean distance and a penalty term for crossing edges along the path. Including gradient penalty term allows the geodesic paths to bend around image gradients to reduce the number of star centers needed to cover a complex shaped object. However, it also makes the star center locations tricky to find as Fig.\ref{fig:geodesicPath} shows an intuitive placement of star centers leading to failed segmentation (discussed more in Sec.\ref{sec:discussion}). }

If more detailed template or pre-segmentation is available, then the distance transform of the aligned template can be incorporated into the pairwise term of the graph-cut framework to softly penalize the difference between segmentation and the template~\cite{Vu2008,Aslan2010}. This method, however, requires a balance between the soft shape prior term and the other image information terms, which may be hard to tune. 



\subsection{Overview of Contributions}
A gradient vector flow (GVF)~\cite{Xu1998} defines a diffusion of the gradient vectors from a grayscale or binary edge map. When computed from an edge map, it defines a vector roughly pointing towards the closest strong edge at each voxel. In this paper, we show that if computed from a volumetric pre-segmentation, instead of an edge map, the GVF encodes descriptive shape information from the pre-segmentation. 
%
%
We thus propose a novel GVF-based shape prior which can be \emph{directly} embedded  into the voxel grid space. This shape prior representation avoids the cumbersome self-intersection detection problem in the mesh representation. It   enables a simple incorporation of the inclusion/exclusion relationship in segmenting multiple interacting objects with shape priors. The segmentation is formulated as a Markov random field (MRF) optimization problem encoded within it the GVF-based shape prior. It turns out that the minimum $s$-$t$ cut algorithm can be used to solve the MRF optimization problem to achieve a globally optimal solution. 

\section{Methods}

We first introduce the gradient vector flows and the GVF-based shape priors, then we show how to incorporate such a shape prior into an MRF formulation and how to segment multiple interacting objects simultaneously. 

\subsection{Gradient Vector Flows}

\begin{figure*}
	\centering
	\subfloat[Gradient]{\includegraphics[width=0.3\linewidth]{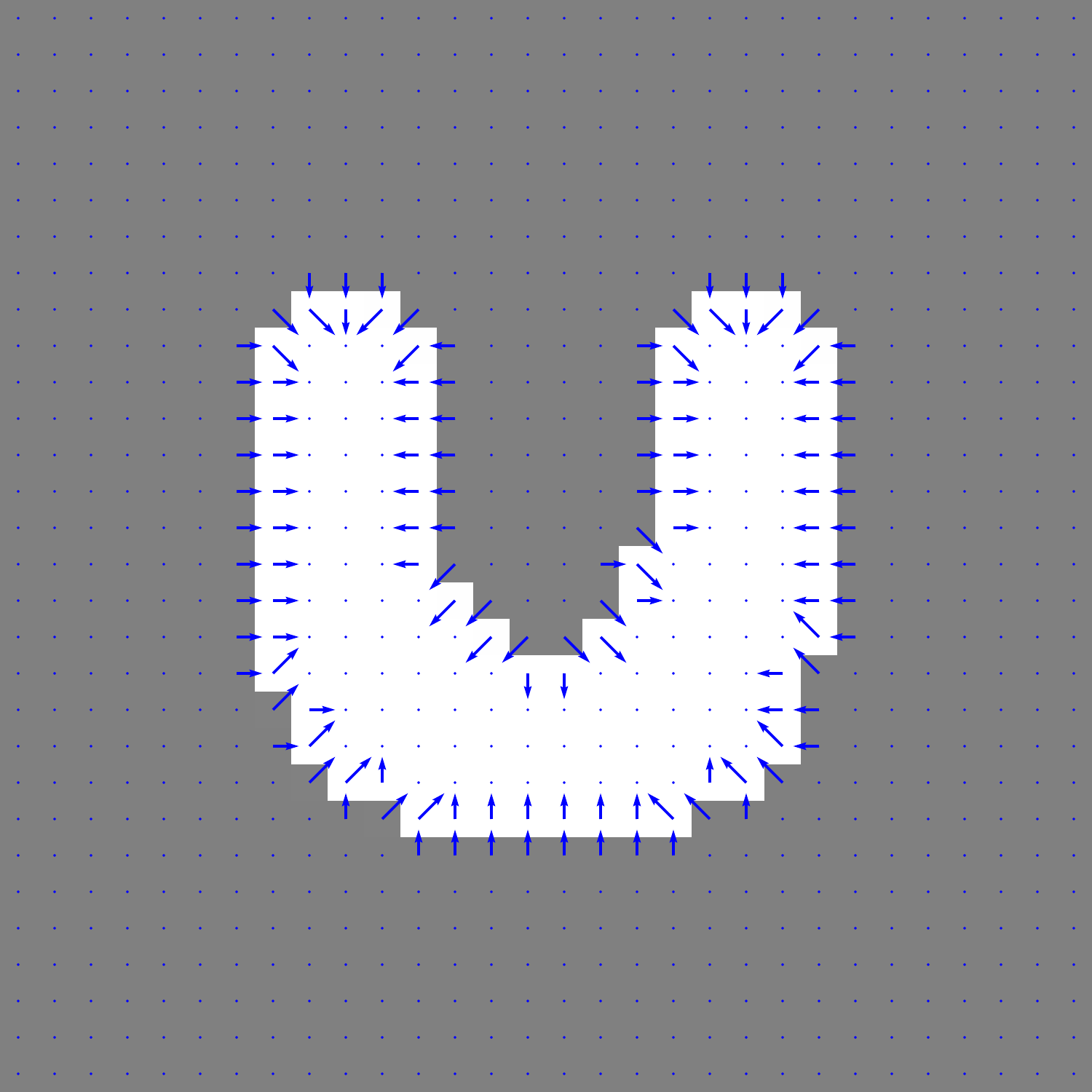}%
		\label{subfig:gradient}}
	\hfill
	\subfloat[GVF]{\includegraphics[width=0.3\linewidth]{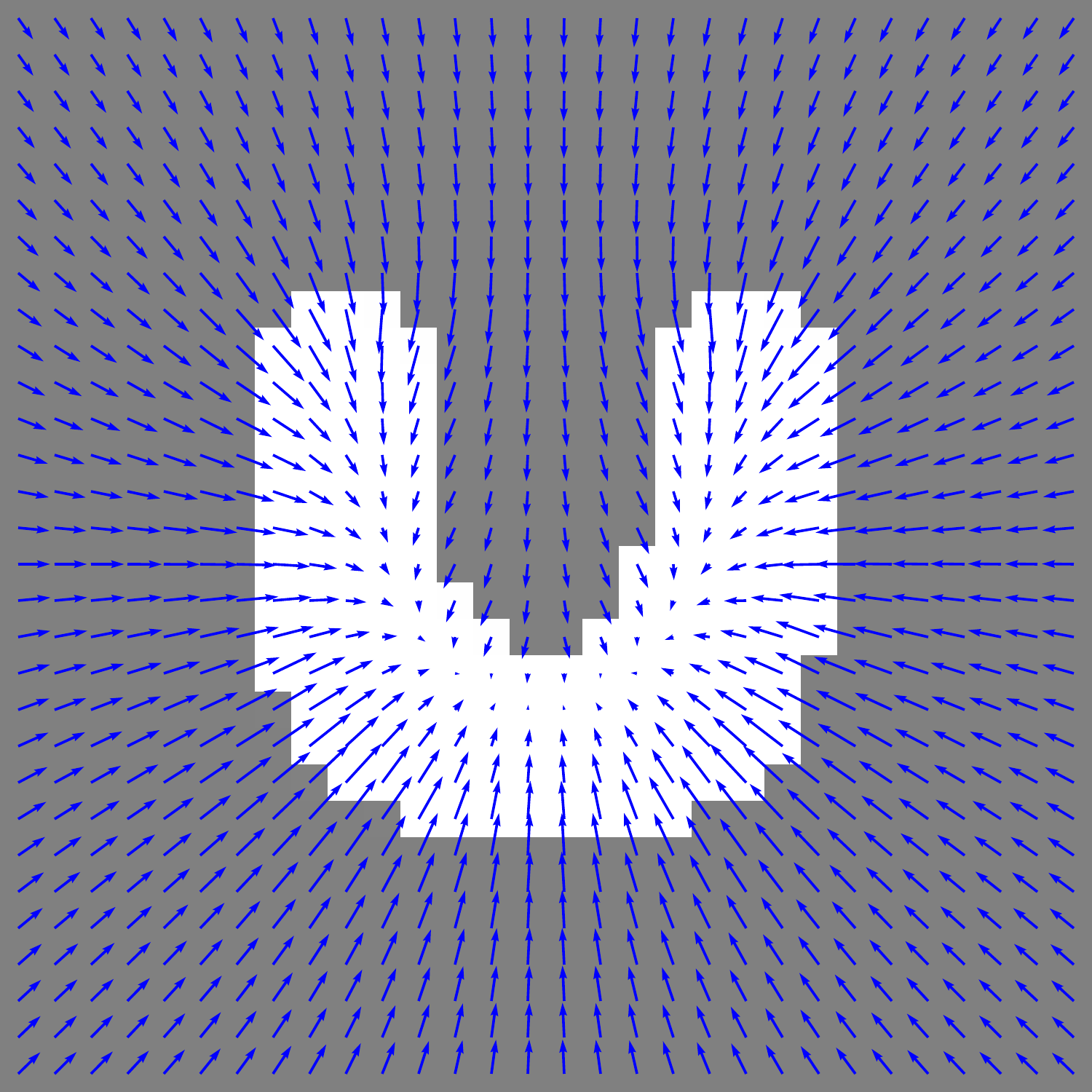}%
		\label{subfig:GVF}}
	\hfill
	\subfloat[discretized GVF]{\includegraphics[width=0.3\linewidth]{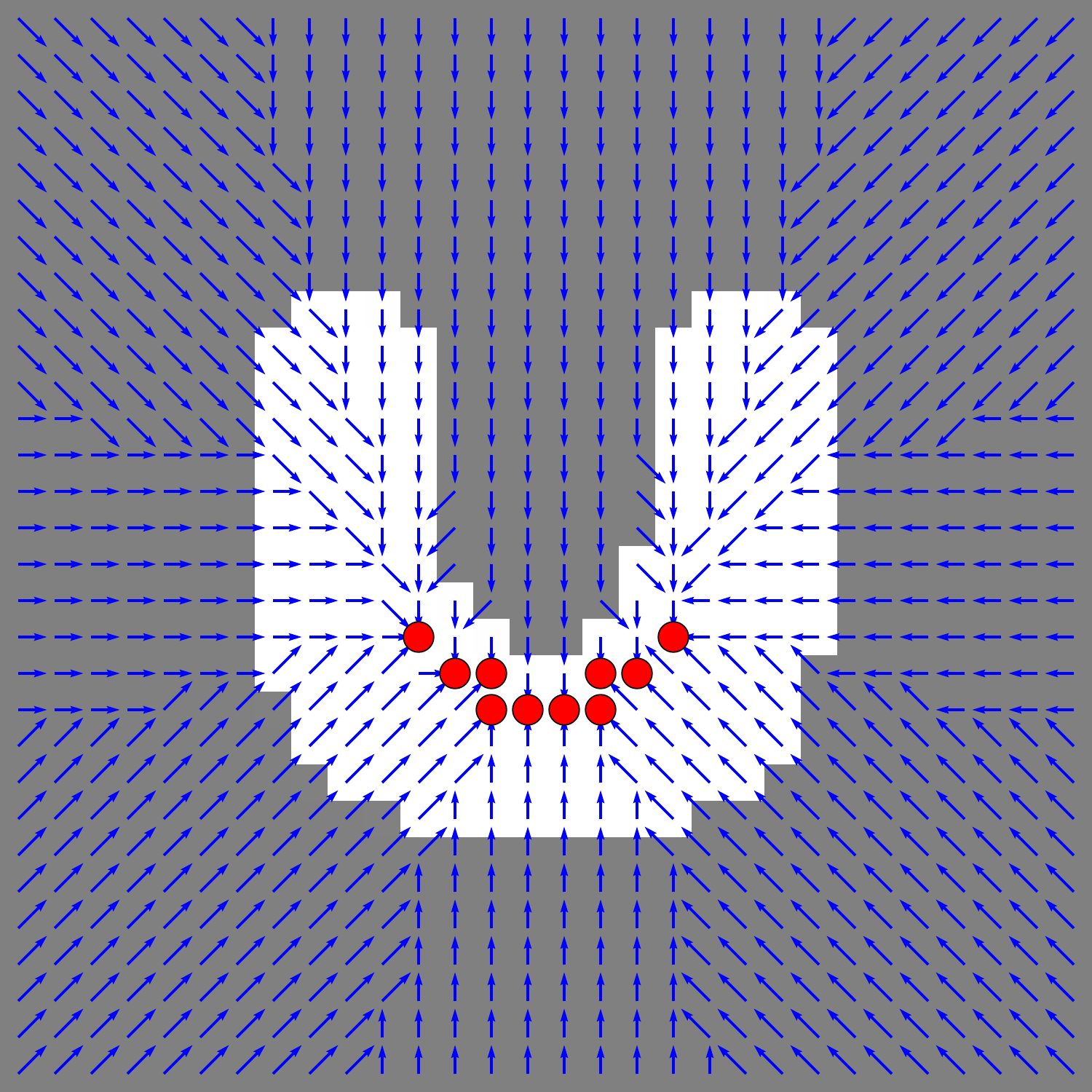}%
		\label{subfig:discretizedGVF}}
	\caption{The use of GVF to encode the shape prior information in the image. \protect\subref{subfig:gradient} shows that the raw gradient of the pre-segmentation is restricted to a narrow band around the boundary. GVF can propagate it smoothly over the whole image, as in \protect\subref{subfig:GVF}. A discretized GVF in \protect\subref{subfig:discretizedGVF} generates a GVF path for every non-core voxels connecting to one of the core voxels (red circles).}
	\label{fig:GVFBuilding}
\end{figure*}

Suppose $\mathcal{P}$ is the set of image voxels in a 3D grid. The binary segmentation assigns every voxel $p\in \mathcal{P}$ a label $f_p \in \mathcal{L}$, where $\mathcal{L}=\{0,1\}$ is the set of available labels. Assume that a pre-segmentation $\hat{S} = \{p|\hat{f}_p = 1, p\in \mathcal{P}\}$ is available, then the gradients of the pre-segmentation $\nabla \hat{S}$ are those nonzeroes around the pre-segmentation boundary (Fig.\ref{subfig:gradient}). In the narrow band with $\nabla \hat{S} \not= 0$, the gradient vectors encode the shape information of the pre-segmentation in the sense that they are the most direct direction pointing towards the inside of the pre-segmentation. 

In order to enforce the shape prior in a larger range, we propose to use GVF~\cite{Xu1998} to propagate the gradient vectors. The GVF takes the raw gradient vector field $\nabla \hat{S}$ as input, and computes a new vector field $\mathbf{h}$ to minimize the following energy
\begin{align}
\mathcal{E}^{GV\!F} (\mathbf{h})= \iiint_{\mathcal{P}} \|\nabla \hat{S}\|^2 \|\mathbf{h} - \nabla \hat{S}\|^2 + \mu \|\nabla \mathbf{h}\|^2 \,dx \,dy \,dz .
\end{align}
The first term is the product of the gradient magnitude and the squared difference of the output vector field and the input gradient. The second term is the sum of the squares of the partial derivatives of the output GVF vectors. More specifically, the first term, which is dominant when the input gradient $\nabla \hat{S}$ is strong, encourages the output vector field $\mathbf{h}$ to well align with the input gradient. \xxx{The second term, which is dominant when the input gradient is small, ensures the output vector field to change smoothly in the regions far away from the region with strong gradients as the input. The constant $\mu$ balances the fidelity to the original input gradient field $\nabla \hat{S}$, and the smooth extrapolation of the original gradients to remote regions.}

\xxx{Suppose we set $\mu$ to be zero, then the output GVF field $\mathbf{h}$ would be exactly the same as the input field $\nabla \hat{S}$. In another word, $\mathbf{h}$ would only be nonzero inside a narrow band around pre-segmentation boundary, and drops immediately to zeros once outside this narrow band. By setting $\mu$ to a nonzero value, the second term would gradually smooth this sudden drop in the output $\mathbf{h}$, effectively propagating the gradient information farther beyond the narrow band. The larger $\mu$ is, the longer the gradient propagation range is. If we set $\mu$ to a very large value, then it can effectively propagate the gradient throughout the whole image. Fig.\ref{subfig:GVF} shows such an example where a GVF vector field propagates the gradient information (Fig.\ref{subfig:gradient}) to all voxels of the image. However, setting $\mu$ to a larger value leads to longer time for computing the GVF field. Thus, one may only need to set $\mu$ to a value that covers the target object boundary search range.}


\subsection{GVF-based Shape Prior in the Voxel Grid Space}
From Fig.\ref{subfig:GVF}, we can see that the GVF magnitude is very small at the ``core'' of pre-segmentation due to the competition of boundary gradients from multiple directions. 
Formally we define the object \emph{``core''} $\mathcal{C}$ as the set of voxels inside the pre-segmentation with GVF magnitude smaller than a small threshold $\theta_\mathcal{C}$, i.e., $\mathcal{C} = \{p \in \mathcal{P}| p \in \hat{S}, \|\mathbf{h}_p \| < \theta_\mathcal{C}\}$. The red circled voxels in Fig.\ref{subfig:discretizedGVF} are the core voxels of the white pre-segmentation. 

For every non-core voxel in the GVF field, we can find a path following which one can travel to one of the core voxels. If one voxel is part of the object, then all voxels along the GVF path connecting it to the core should also be part of the object if we want to roughly preserve the shape of the pre-segmentation. Otherwise, a hole or an undesired cavity may appear along the GVF paths. 

To label each voxel in the voxel grid space, we discretize the GVF field so that the GVF vector at each voxel points  towards exactly one of its neighbor voxels. More concretely, the GVF vector field $\mathbf{h}$ is transformed into the discretized GVF vector field $\mathbf{h}^D$. For every pixel $p$, $\mathbf{h}^D_p = \overrightarrow{pq}$ where $q = \argmax_{q\in \mathcal{N}_p} \mathbf{h}_p \cdot \overrightarrow{pq} / (\|\mathbf{h}_p\| \|\overrightarrow{pq} \|)$. $\mathcal{N}_p$ is the set of neighboring voxels of voxel $p$. In another word, $\mathbf{h}_p^D$ is the vector from $p$ to its neighbor voxel with the smallest angle error when approximating GVF vector $\mathbf{h}_p$. See Fig.\ref{subfig:discretizedGVF} for an example of the discretized GVF field. 

\begin{figure}
	\centering
	\subfloat[GVF-based shape prior]{\includegraphics[width=0.4\linewidth]{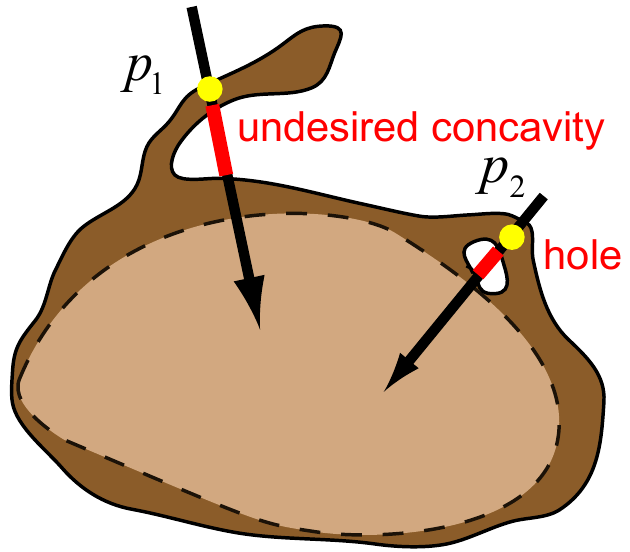}%
		\label{subfig:gvfShapePrior}}
	\hfill
	\subfloat[inclusion]{\includegraphics[width=0.29\linewidth]{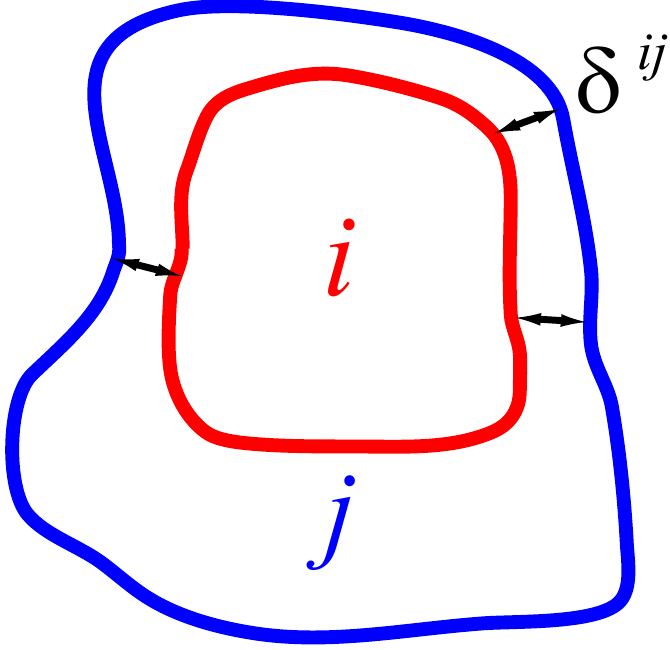}%
		\label{subfig:inclusion}}
	\hfill
	\subfloat[exclusion]{\includegraphics[width=0.29\linewidth]{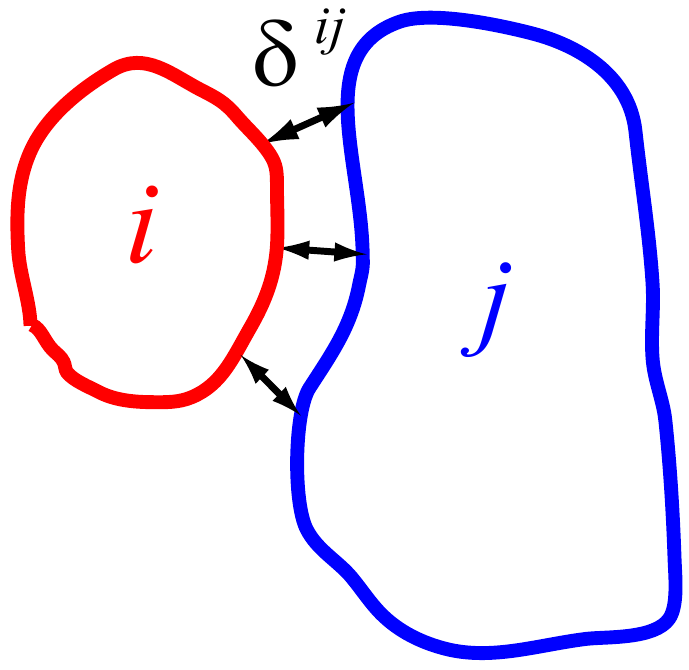}%
		\label{subfig:exclusion}}
	\caption{Illustrating a GVF-based shape prior and the inclusion/exclusion multi-object interaction priors. \protect\subref{subfig:gvfShapePrior} demonstrates that exaggerated concavity and holes not existing in pre-segmentation would violate the GVF-based shape prior. \protect\subref{subfig:inclusion} and \protect\subref{subfig:exclusion} show the inclusion and exclusion interaction relationship between objects $i$ and $j$ with a minimum distance $\delta^{ij}$ in between.}
	\label{fig:priorGeoDemo}
\end{figure}

For any non-core voxel $p \notin \mathcal{C}$, the \emph{GVF path} connecting itself to the object core is defined as $G\!P(p)=q_0, q_1, \ldots, q_N$, such that $\overrightarrow{pq_0}, \overrightarrow{q_0 q_1}, \ldots,\overrightarrow{q_{K-1} q_N} \in \mathbf{h}^D$, and $q_N \in \mathcal{C}$. 
The \emph{GVF-based shape prior} is defined as: if one voxel $p$ is part of the object, then all voxels along the GVF path $G\!P(p)$
are also part of the object. Geometrically, the boundary surface of the target object is monotone with respect to the GVF paths, that is, the surface intersects with any GVF path at most once. Thus, we maintain the global structure of the segmented object to align to the shape of the pre-segmentation or the initial model.
For instance, in Fig.\ref{subfig:gvfShapePrior}, while voxel $p_1$ ($p_2$) is in the foreground, the  voxels on part of the GVF path connecting $p_1$ ($p_2$) to the object core (the thick black lines) are not labeled as the foreground (highlighted by thick red lines). The GVF-based shape prior computed from the pre-segmentation (the light brown region) is thus violated by the current segmentation (the deep brown region), due to the undesired concavity and the hole. 

\subsection{Penalty Functions for the GVF-based Shape Prior}

For a single voxel $p$, the GVF-based shape prior can be enforced by the following penalty function between $p$ and all $q \in G\!P(p)$.
\begin{align}
\phi (f_p, f_q) = \infty \cdot [f_p=1, f_q=0],
\label{eq:gvfPhiRaw}
\end{align}
where $[\cdot]$ is the indicator function which returns $1$ when the enclosed condition is true, and 0 otherwise. The penalty function $\phi(f_p,f_q)$ returns $\infty$ when $p$ belongs to the object, while $q$ is in the background. Otherwise, the penalty function returns $0$. In a minimization problem, this ensures that any solution violating the GVF-based shape prior is not valid. 

Due to the possible large number of voxels along the GVF path, enforcing $\phi (f_p,f_q)$ between $p$ and all $q \in G\!P(p)$ naively is computationally expensive, especially considering that this process has to be repeated for every voxel $p \in \mathcal{P}$. Fortunately, it can be shown that the GVF-based shape prior (i.e., the monotonicity of the boundary surface of the target object with respect to all the GVF paths)  can be sufficiently enforced by using only $\phi(f_p,f_{q_0}), \phi(f_{q_0}, f_{q_1}), \ldots, \phi(f_{q_{K-1}}, f_{q_N})$ where $q_0, q_1, \ldots, q_N = G\!P(p)$~\cite{Wu2002,Veksler2008}. Thus, the penalty function only need to be enforced between $p$ and its nearest neighbor indicated by the discretized GVF vector $\mathbf{h}_p^D$. This leads to the observation that to enforce the GVF-based shape prior for all voxels $p \in \mathcal{P}$, we only need to iterate through the discretized GVF vector field $\mathbf{h}^D$ ($\left\vert\mathbf{h}^D\right\vert = \left\vert\mathcal{P} - \mathcal{C}\right\vert$), i.e., 
\begin{align}
\sum_{\overrightarrow{pq}\in \mathbf{h}^D} \phi(f_p,f_q) = \sum_{\overrightarrow{pq}\in \mathbf{h}^D}{\infty \cdot [f_p=1, f_q=0]}.
\label{eq:gvfPhi}
\end{align}

\subsection{MRF Formulation}

Suppose $\mathcal{N}_\mathcal{P}$ is the neighborhood system in image $\mathcal{P}$, then the overall MRF energy with the GVF-based shape prior for segmentation is
\begin{align}
\mathcal{E}(\mathbf{f}_{\mathcal{P}}) = \sum_{p \in \mathcal{P}} D(f_p) + \sum_{(p,q) \in \mathcal{N}_\mathcal{P}}{V_{pq}(f_p,f_q)} + \sum_{\overrightarrow{pq}\in \mathbf{h}^D} \phi(f_p,f_q).
\label{eq:mrf}
\end{align}

$D_p(f_p)$ is a data term describing the appearance information of voxel $p$. The more likely it belongs to the foreground, the smaller $D_p(f_p=1)$ is, while the larger $D_p(f_p=0)$ is. $V_{pq}(f_p, f_q)$ is a pairwise smoothness term based on the boundary/edge information between neighboring voxels $p$ and $q$. If $f_p=f_q$, then $V_{pq}(f_p, f_q) = 0$; otherwise, $V_{pq}(f_p, f_q)$ is a nonnegative number that is inversely proportional to the likelihood of an edge between the neighboring voxels $p$ and $q$. In general, the smoothness term encourages a smooth boundary between the object and the background. The more likely an edge lies in between the neighboring voxels, the smaller the smoothness penalty is. 
The third term in Eq.\eqref{eq:mrf} enforces the GVF-based shape prior as in Eq.\eqref{eq:gvfPhi}. 


\subsection{Solving MRF by minimum $s$-excess problem}
\label{sec-s-excess}
Now we show how to encode the MRF formulation with the GVF-based shape prior as a \emph{minimum $s$-excess problem}, which can be exactly solved by computing a minimum $s$-$t$ cut~\cite{Wu2002,hochbaum_efficient_2001}.
The minimum $s$-excess problem in a directed graph $G=(V,E)$ is defined as a partition of the graph vertices into two disjoint sets according to the sum of weights on specific vertices and edges. More specifically, suppose every vertex $v\in V$ carries an arbitrary weight $w(v)$ and every directed edge $e\in E$ carries a nonnegative weight $c(e)\geq 0$, then the minimum $s$-excess problem seeks a vertex subset $H \subseteq V$ such that the cost $\gamma(H)$ is minimized, with $\gamma(H) = \sum_{v\in H}w(v) + \sum_{\substack{(u,v)\in E\\ u \in H, v\in \bar{H}}}c(u,v)$, where $\bar{H}=V-H$. In another word, the $s$-excess cost of a vertex set (also called \emph{source set}) $H$ is defined as the sum of vertex weights for all vertices inside the set, and the edge weights for all edges from inside the set to outside the set. 

We construct a graph such that when the minimum $s$-excess problem is solved, the voxels corresponding to the vertices in the source set $H$ will be labeled as foreground. Each voxel $p$ in the image defines exactly one vertex $v_p$ in the graph. The three terms in Eq.\eqref{eq:mrf} are correctly encoded in the graph by the following means: 
\begin{itemize}
	\item The data term is encoded by assigning vertex weight $w(v_p) = D_p(1)-D_p(0)$, i.e., the data term difference when being labeled as foreground from background. In this way, the sum of the vertex weights in a source set $H$ is $\sum_{v_p \in H}w(v_p) = \sum_{v_p \in H}(D_p(1)-D_p(0))=\sum_{v_p\in H}D_p(1) + \sum_{v_p \in \bar{H}}D_p(0) - \sum_{p\in \mathcal{P}}D_p(0)$. Note the first two terms are exactly the data term in Eq.\eqref{eq:mrf} and the third term is a constant. 
	
	\item The smoothness term is encoded by adding edges $(v_p,v_q), \forall (p,q)\in \mathcal{N}_{\mathcal{P}}$ and assigning edge weights $c(v_p, v_q)=V_{pq}(1,0)$. Thus, the sum of the ``excess'' edge weights (i.e., edges from source set to outside) will be $\sum_{v_p\in H, v_q \in \bar{H}}c(v_p,v_q) = \sum_{v_p \in H, v_q\in \bar{H}}V_{pq}(1,0) = \sum_{(p,q)\in \mathcal{N}_{\mathcal{P}}}V_{pq}(f_p,f_q)$, because all neighboring voxel pairs have the same label, except the ones with corresponding vertices $v_p\in H$ and $v_q \in\bar{H}$. 
	
	\item The GVF-based shape prior term is enforced by adding edges $(v_p,v_q),  \overrightarrow{pq}\in \mathbf{h}^D$ with an infinite edge weight. Similar to the edges for the smoothness term, it is easy to show that these edges correctly encodes the GVF-based shape prior term in Eq.\eqref{eq:gvfPhi}. Any minimal $s$-excess solution would avoid including these edges as ``excess'' edges due to the associated infinite edges weight, thus enforcing the GVF-based shape prior. 
\end{itemize}

Computing the minimum $s$-excess problem in the constructed graph will return a source set $H$, which corresponds to voxels segmented as foreground. All other voxels are labeled as background.

\subsection{Simultaneous Multi-Objects Segmentation}
\label{subsec:multiObjectSeg}
Medical image applications often require \emph{multiple} objects to be segmented simultaneously, e.g., prostate and bladder~\cite{Song2009,Song2010miccai}, white matter and gray matter in brain~\cite{Oguz2014}. It is advantageous to incorporate the interaction prior information between multiple objects when possible. 
Fig.\ref{subfig:inclusion} and \ref{subfig:exclusion} show examples of inclusion and exclusion interactions between two objects $i$ and $j$, with a minimum distance $\delta^{ij}$ between them. 

To simultaneously segment $K$ interactive objects, we need to determine $K$ binary variables $f_p^1, f_p^2, \ldots, f_p^K \in \{0,1\}$ for each voxel $p$, with $\sum_{k=1}^K f_p^k \leq 1$. Here, $f_p^k=1$ indicates voxel $p$ is part of the $k$-th object, and vice versa. The energy function $\mathcal{E}(\mathbf{f}_\mathcal{P})$ is then defined as $\sum_{k=1}^K \mathcal{E}(\mathbf{f}_\mathcal{P}^k)$.  $\mathcal{E}(\mathbf{f}_\mathcal{P}^k)$ is the energy function for segmenting object $O^k$, defined as same as in Eq.\eqref{eq:mrf}. 

\xxx{We build $K$ subgraphs $G^k=(V^k, E^k), k=1,\ldots,K$ to represent the $K$ binary variables for each voxel. In every subgraph $G^k$, a vertex $v^k_p$ is created for every voxel $p$. The idea is that within every subgraph, the graph is built in a way similar to Sec.\ref{sec-s-excess} to encode the region term, boundary term and the GVF shape prior. Since our proposed GVF shape prior is embedded directly in the voxel space, we can utilize the method proposed by Delong \textit{et al.}~\cite{Delong2009} to add additional edges between two subgraphs corresponding to two interacting objects to enforce the minimum separation distance constraints. }

\begin{figure}
	\centering
	\subfloat[\xxx{multi-object inclusion}]{\includegraphics[width=0.3\linewidth]{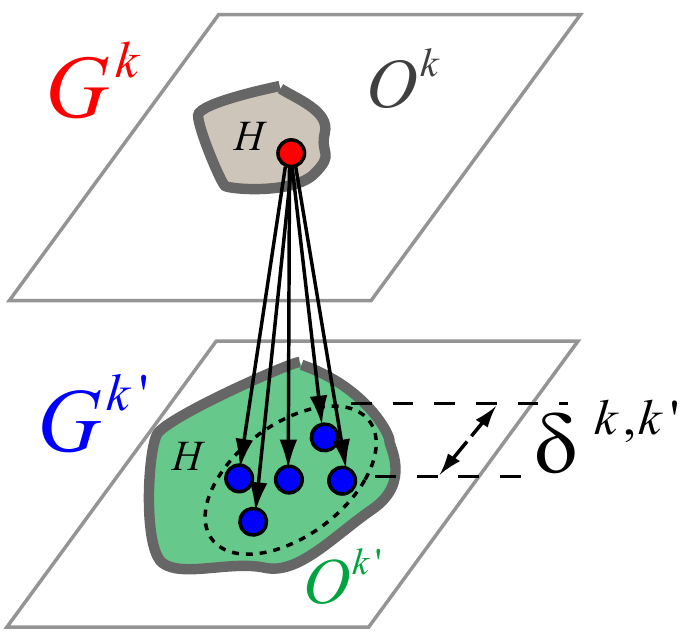}%
		\label{subfig:MOinclusion}}
	\hfill
	\subfloat[\xxx{multi-object exclusion}]{\includegraphics[width=0.3\linewidth]{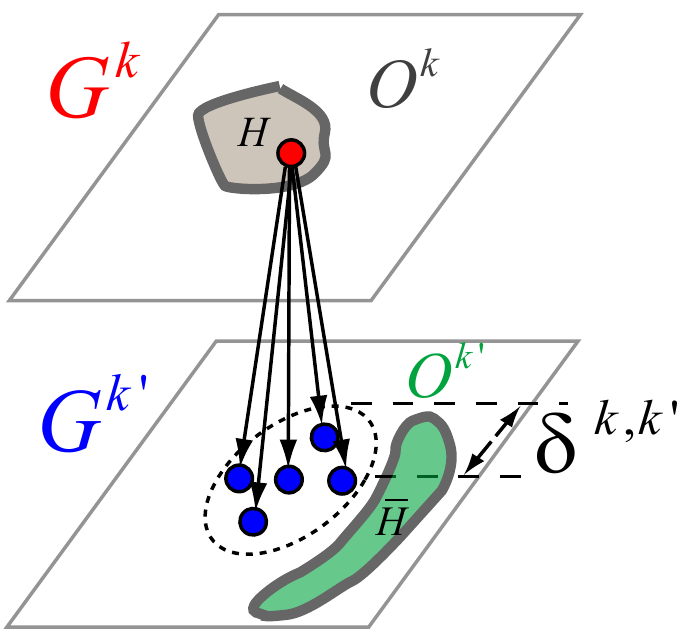}%
		\label{subfig:MOexclusion}}
	\hfill
	\subfloat[\xxx{max distance separation}]{\includegraphics[width=0.3\linewidth]{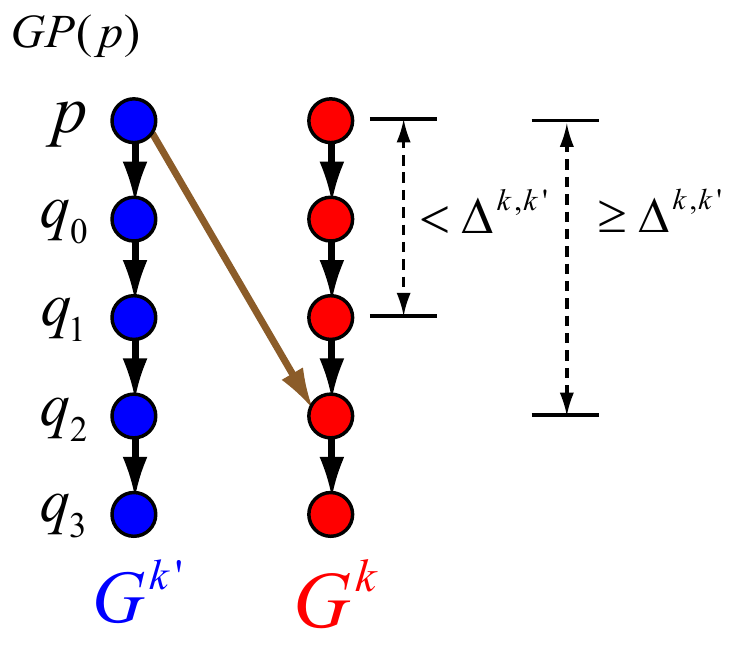}%
		\label{subfig:MOinclusionMaxDist}}
	\caption{\xxx{Illustration of how to enforce the minimum surface separation constraints for the inclusion and the exclusion interaction relations, and how to enforce the maximum surface distance constraints when both objects share the same pre-segmentation. \protect\subref{subfig:MOinclusion} and \protect\subref{subfig:MOexclusion} show that additional edges within a ``cone'' of radius $\delta^{k,k'}$ need to be added to enforce the minimum distance between the nested and the excluded objects, respectively. \protect\subref{subfig:MOinclusionMaxDist} shows the added edges between subgraphs for enforcing the maximum distance between two object surfaces sharing the same pre-segmentation and thus using the same GVF paths.} }
	\label{fig:MO}
\end{figure}

\xxx{Now we discuss inclusion and exclusion scenarios separately since they differ slightly in graph construction. Note although we define interaction relationship only between two objects at a time, we can segment multiple interacting objects simultaneously since one object can interact with multiple other objects.}

\xxx{\paragraph{Inclusion/nesting relationship} Suppose object $O^k$ is inside object $O^{k'}$ with a minimum distance of $\delta^{k,k'}$ between their boundaries. Without loss of generality, we assume $v_p^k \in H \iff f_p^k=1$, i.e., the vertex for pixel $p$ in subgraph $G^k$ belonging to the source set in the solution to the $s$-excess problem corresponds to labeling pixel $p$ as part of the $k$-th object $O^k$ (as presented in Sec.\ref{sec-s-excess}). For both subgraphs $G^k$ and $G^{k'}$, the vertex weights for the region term, the edge weights for the boundary term, and the edge weights for the GVF shape prior are assigned exactly the same as presented in Sec.\ref{sec-s-excess}.}

\xxx{To enforce the minimum separation constraints, we add edges $(v_p^k, v_{p'}^{k'})$ with an $+\infty$ weight for all voxels $p'$ such that $\|\overrightarrow{pp'}\|\leq \delta^{k,k'}$. As shown in Fig.\ref{subfig:MOinclusion}, these edges are from the vertex for voxel $p$ in subgraph $G^k$ to all vertices within the minimum distance from the voxel $p$ in subgraph $G^{k'}$. The rationale is that if vertex $v^k_p$ is in the source set, then vertex $v^{k'}_{p'}$ must also be in the source set. In another word, if voxel $p$ is part of the $k$-th object, then all voxels $p'$ that are ``close" to it (closer than $\delta^{k,k'}$) must also be part of the $k'$-th object when we assume $v_p^{k'} \in H \iff f_p^{k'}=1$. This ensures that the boundary of the $k$-th object is always inside the $k'$-th object with a minimum distance of $\delta^{k,k'}$ between them. }

\xxx{\paragraph{Exclusion relationship} Suppose object $O^k$ does not overlap with object $O^{k'}$ with a minimum distance of $\delta^{k,k'}$ in between. Without loss of generality, we assume $v_p^k\in H \iff f_p^k=1$, i.e., the vertex for pixel $p$ in subgraph $G^k$ belonging to the source set in the solution to the $s$-excess problem corresponds to labeling pixel $p$ as part of the $k$-th object $O^k$. 
For subgraph $G^k$, the vertex weights for the region term, the edges and edge weights for the boundary term and GVF shape priors, are the same as presented in Sec.\ref{sec-s-excess}. For subgraph $G^{k'}$, the vertex weights for the region term and the edges and edge weights for the boundary term are also the same as presented in Sec.\ref{sec-s-excess}. The edge weights between the two subgraphs for the minimum separation distance constraints are also assigned in the same way as for the inclusion relationship. }

\xxx{However, we do need to flip the meaning of vertices in subgraph $G^{k'}$ such  that $v_p^{k'} \in H \iff f_p^{k'}=0 $. Correspondingly, the edge directions for the GVF shape priors in subgraph $G^{k'}$ need to be reversed.}

\xxx{First, we analyze why flipping the meaning of $v_p^{k'}$ and adding edges $(v_p^k, v_p^{k'})$, $\|\overrightarrow{pp'}\|\leq \delta^{k,k'}$ with $+\infty$ weights correctly enforces the minimum distance separation constraints. 
Referring to Fig.\ref{subfig:MOexclusion}, if voxel $p$ is part of the $k$-th object, then vertex $v_p^k$ is in the source set in the solution to the $s$-excess problem. The edges with an infinite weight ensure that vertex $v_{p'}^{k'}$ with $\|\overrightarrow{pp'}\| \leq \delta^{k,k'}$ also belongs to the source set. As we flip the meaning of vertices in subgraph $G^{k'}$, that is, a vertex in the source set corresponds to the voxel being labeled as \emph{not} part of the $k'$-th object, any voxel that is within $\delta^{k,k'}$ from object $O^k$ are not part of $O^{k'}$. This enforces the exclusion relation between the two objects with a minimum separation distance. }

\xxx{Now that we flip the mapping from vertices $v_{p'}^{k'}$ to labeling variables $f_{p'}^{k'}$, we have to reverse the directions of the edges for the GVF shape priors in subgraph $G^{k'}$ in order to encode the same GVF shape prior energy term in Eq.\eqref{eq:gvfPhi}. }

It is also possible to enforce the \emph{maximum} surface distance between pairs of nested surfaces when they share the same pre-segmentation. In this case, the two surfaces share the common GVF paths which serve as graph columns in the grid space. This common column structure enables the maximum surface distance constraint to be enforced using techniques used in LOGISMOS~\cite{Bai2014}. 

\xxx{More specifically, suppose object $O^k$ is inside $O^{k'}$ and the two object boundaries are at most $\Delta^{k,k'}$ distance apart. Assume $GP(p)=q_0, q_1, \ldots, q_N$ is the shared GVF path connecting voxel $p$ to the object core for both objects $O^k$ and $O^{k'}$, we add an edge $(v_p^{k'}, v_{q_i}^k)$ such that $q_i \in GP(p)$ with $\|\overrightarrow{pq_i}\| \geq \Delta^{k,k'}$ and  $\|\overrightarrow{pq_{i-1}}\| < \Delta^{k,k'}$, i.e., $q_i$ is the first voxel along the GVF path that is more than $\Delta^{k,k'}$ away from voxel $p$. This ensures that if voxel $p$ is in object $O^{k'}$, then all voxels that are at least $\Delta^{k,k'}$ away along the GVF path must also be in object $O^k$. In another word, the boundary distance between the two objects are never greater than the maximum separation distance constraints. Fig.\ref{subfig:MOinclusionMaxDist} shows how edges are added to enforce the maximum distance constraints while nested surfaces share the same pre-segmentation. }

\section{Experiment}
We validated the proposed method on two applications: the brain tissue segmentation in MRI T1 images, and the prostate and bladder segmentation in CT images. The brain tissue segmentation aims to segment two \emph{nested} surfaces which separates three different types of tissues (white matter, gray matter, and cerebrospinal fluid), while the second application aims to segment two \emph{mutually exclusive} objects. 

\subsection{Experiment Settings}

For both applications, we first obtain an initial segmentation, i.e., pre-segmentation. The proposed GVF shape prior is defined based on this pre-segmentation. The final segmentation output accuracy is assessed by two metrics: Dice similarity coefficient (DSC) and average symmetric surface distance (ASSD). The \emph{Dice similarity coefficient} is used to measure how well two volumes overlap with each other. Assume $A$ and $B$ are two volumes, then the DSC between the two volumes is defined as $2|A\cap B|/(|A|+|B|)$, which ranges between $0$ and $1$.  The larger the DSC is, the better the two volumes are aligned, with $1$ indicating a perfect overlapping. The \emph{average symmetric surface distance} (ASSD) is used to measure how close two segmented surfaces are. Assume given segmentation volume $A$, boundary surface $S_A$ is defined to be the set of voxels in $A$ that has at least one neighboring background voxel. Let $d(x,S_A)$ denote the shortest distance between a point $x$ and any point on the boundary surface $S_A$ of segmentation $A$. The ASSD between two segmentations $A$ and $B$ is defined as $ASSD=({\sum_{a\in S_A}d(a,S_B) + \sum_{b\in S_B}d(b,S_A)})/{(|S_A|+|S_B|)}$. It measures the average distance from any point on a contour/surface to the other contour/surface. The ASSD metric has a range of $[0,+\infty]$. The smaller ASSD is, the better the two segmented surfaces/contours agree with each other. If $ASSD=0$, then the two segmentations are identical. 

\subsection{Brain Segmentation: Data and Compared Methods}
\label{subsec:brainDataCompareMethods}
For the brain tissue segmentation, 18 T1-weighted scans of normal subjects from the Internet Brain Segmentation Repository (IBSR)~\cite{IBSR} were used (commonly know as ``IBSR 18'' in the literature)~\cite{Wels2011,Valverde2015}. The image size is $256\times 128\times 256$ with the voxel spacing range from $0.837\times 1.5 \times 0.837 mm^3$ to $1\times 1.5 \times 1 mm^3$. Each scan comes with a brain mask marking voxels inside skull. Each voxel inside the brain mask is labeled by expert as one of three labels: white matter (WM), gray matter (GM), and cerebrospinal fluid (CSF). Using this public dataset allows us to directly compare the proposed method to various state-of-art methods. 

Valverde \textit{et al.}~\cite{Valverde2015} validated 10 brain segmentation algorithms on the IBSR dataset, aiming to evaluate a wide set of available techniques and tools. We compare our method to the results reported. These ten algorithms are summarized below: 
FAST~\cite{Zhang2001} uses expectation maximization with K-means initialization to optimize an MRF model; 
SPM5~\cite{Ashburner2005} and SPM8~\cite{ashburner2012spm8} are two versions of the SPM toolbox based on the iterative Gaussian mixture model, atlas registration and bias correction techniques; 
GAMIXTURE~\cite{Tohka2007} estimates a Gaussian mixture model by a genetic algorithm; 
ANN~\cite{Tian2007} implements a self organizing map for clustering image data; 
FCM~\cite{Pham2001} uses a fuzzy c-means clustering algorithm; 
KNN~\cite{DeBoer2009} performs $k$-nearest neighbor searches based on an automated registration of prior probability atlases; 
SVPASEG~\cite{Tohka2010} applies an iterative conditional modes (ICM) method with an initialization based on a genetic algorithm to optimize an MRF model;  
FANTASM~\cite{Pham} extends FCM by adding a spatial term in the objective function; 
and PVC~\cite{Shattuck2001} builds a maximum-a-posteriori (MAP) model with the ICM optimization. 
We refer readers to Table.1 in~\cite{Valverde2015} for software sources and versions used in reporting these numbers. 

In addition to the above 10 methods, we also included the results reported by the following state-of-art segmentation methods: (i) the discriminative model-constrained EM approach, which combines supervised discriminative modeling and unsupervised statistical expectation maximization (EM) segmentation into an integrated Baysian framework~\cite{Wels2011}; (ii) the adaptive Markov modeling based on mutual information~\cite{Awate2006}; (iii) the hidden Markov chain model~\cite{Bricq2008} which unifies partial volume effect, bias field correction, and a probabilistic atlas; and (iv) the prior knowledge driven multiscale segmentation~\cite{Akselrod-Ballin2007} which embeds an atlas prior in a multi-scale pyramid. 

\xxx{We also compared our method to Atropos~\cite{Avants2011}, the segmentation tool distributed with the state-of-art brain MRI registration software Advanced Normalization Tools (ANTs)~\cite{Rajchl2016,Ou2014,Klein2009}. It solves $n$-class segmentation formulated as a Baysian optimization problem by an expectation maximization (EM) method with Markov random field (MRF) modeling. We used the software downloaded from the software author's website~\cite{ANTs}, with the same pre-segmentation used in our proposed method (Sec.\ref{subsubsec:preseg}) as the initial labeling for EM iterations, The MRF modeling smoothing parameter was set to 0.2 with every voxel as an MRF variable after a grid search for optimal parameters. }



\subsection{Brain Segmentation: Workflow}
\label{subsec:brainWorkflow}
\subsubsection{Preprocessing} Each image is first \xxx{cropped by the brain mask to only include voxels inside the skull}, 
reoriented to have the standard orientation (`RAI' orientation), and resampled to have unit voxel spacing. We use 2-fold cross validation for this experiment. The 18 images are randomly split into two disjoint set, with one of them being training set and another being test set. Then the roles of training/testing set are reversed for both sets. To account for the inter-scan intensity profile difference, we first compute a mean brain histogram of the training set. Then we run histogram matching to match every test image to the histogram. 

\subsubsection{Pre-segmentation} 
\label{subsubsec:preseg}
The BRAINSABC software\footnote{Available online: \url{https://github.com/BRAINSia/BRAINSTools}}~\cite{YoungKim2013} is used to classify every voxel within the brain mask into one of the three tissue classes: WM, GM, and CSF. BRAINSABC first deformably registers the input image to an atlas, then performs atlas-based tissue classification using an expectation-maximization approach~\cite{VanLeemput1999}. The largest connected component consisting of white matter voxels is used as the pre-segmentation of the inner surface, i.e., WM-GM boundary. The largest connected component consisting of white matter \emph{and} gray matter voxels is used as the pre-segmentation of the outer surface, i.e., GM-CSF boundary. The GVF-based shape priors are defined with respect to these pre-segmentations. To simplify notation, we would call the WM-GM boundary as the WM surface, and the GM-CSF boundary as the GM surface. The WM surface is set to be included in the GM surface with a minimum distance of 1 $mm$. 

\subsubsection{Energy term design} The random forest method using a simple four-dimensional feature vector is  applied to generate voxel-wise probability maps for all three types of tissues. The four features are the \xxx{\emph{normalized}} $x, y, z$ coordinates \xxx{(normalized to [0,1] range)} and the intensity of each voxel. Using the techniques in Section~\ref{subsec:multiObjectSeg}, two variables $f_p^\text{WM}, f_p^\text{WM/GM} \in \{0,1\}$ are introduced for each voxel $p$. \xxx{If $f_p^{WM}=1$, then $p$ is WM. If $f_p^{WM/GM}=1$, then $p$ is WM \emph{or} GM.} The corresponding data terms $D(\cdot)$ in Eq.\eqref{eq:mrf} for the two labels are defined in Eq.\eqref{eq:brainDataTermPos} and \eqref{eq:brainDataTermNeg}. \xxx{Fig.\ref{subfig:brainCostRegionTerm} shows an example WM region term $D^{WM}(f^{WM}=1)$ computed from input image Fig.\ref{subfig:brainCostInput}. }
\begin{align}
\begin{split}
D_p^\text{WM}(f_p^\text{WM}=1)  &\propto - \log \Pr (\text{WM}|p), \\
D_p^\text{WM/GM}(f_p^\text{WM/GM}=1)  &\propto - \log \Pr (\text{WM/GM}|p).
\end{split}
\label{eq:brainDataTermPos}
\end{align}
\begin{align}
D_p^\text{l}(f_p^l = 0)  \propto -\log(1- \Pr (l|p)), l \in \{\text{WM}, \text{WM/GM}\}.
\label{eq:brainDataTermNeg}
\end{align}

Note that the two surfaces WM-GM and GM-CSF are highlighted at different intensity ranges. It makes sense to adjust the contrast of images when computing the boundary terms $V(\cdot)$ in Eq.\eqref{eq:mrf}. More specifically, the intensity is transformed by a sigmoid function controlled by parameters $\alpha^l$ and $\beta^l$ in Eq.\eqref{eq:sigmoid}, which enhances the contrast roughly within the range $[\beta^l-3\alpha^l, \beta^l + 3 \alpha^l]$. Here $l \in \{\text{WM}, \text{GM/WM}\}$. These parameters are set to $(\alpha^\text{WM}, \beta^\text{WM})=(10, 180)$ and $(\alpha^\text{GM/WM}, \beta^\text{GM/WM})=(30, 120)$ by visually checking the training set images. Since the histograms of all images are matched to the reference mean histogram of the training datasets, the contrast adjustment parameters above should be robust to different scans. \xxx{Fig.\ref{subfig:brainCostSigmoid} shows the result of contrast adjustment before computing the WM-GM boundary terms. }

\begin{align}
\bar{I}_p^l & \propto \frac{1}{1+\exp(-\frac{I_p-\beta^l}{\alpha^l})}. \label{eq:sigmoid}\\
V_{pq}(f_p^l \neq f_q^l) & \propto \exp (-(\bar{I}^l_p - \bar{I}^l_q)/\sigma^2), \label{eq:brainBoundaryTerm}\\
V_{pq}(f_p^l=f_q^l) & = 0.
\end{align}

The gradient between neighboring voxels $p$ and $q$ is then used as the boundary term (Eq.\eqref{eq:brainBoundaryTerm}). The more intensity difference of the neighboring voxels has, the smaller the boundary penalty is. On the other hand, if the neighboring voxels have very similar intensities, then the boundary penalty will be large, strongly encouraging the neighboring voxels to share the same labeling. The parameter $\sigma$ in Eq.\eqref{eq:brainBoundaryTerm} is set to 0.1 based on the training set scans. \xxx{Fig.\ref{subfig:brainCostBndTermVertical},\ref{subfig:brainCostBndTermHorizontal} show the boundary terms for vertical and horizontal edges, computed from the contrast-adjusted image (Fig.\ref{subfig:brainCostSigmoid}). }

\subsection{Brain Segmentation: Results}

\begin{figure*}
	\centering 
	\begin{tabular}{>{\centering\arraybackslash}m{.02\linewidth}>{\centering\arraybackslash}m{.28\linewidth}>{\centering\arraybackslash}m{.28\linewidth}>{\centering\arraybackslash}m{.28\linewidth}}
		& original image & manual & proposed \\
		\rotatebox{90}{sagittal view} &
		\includegraphics[width=\linewidth]{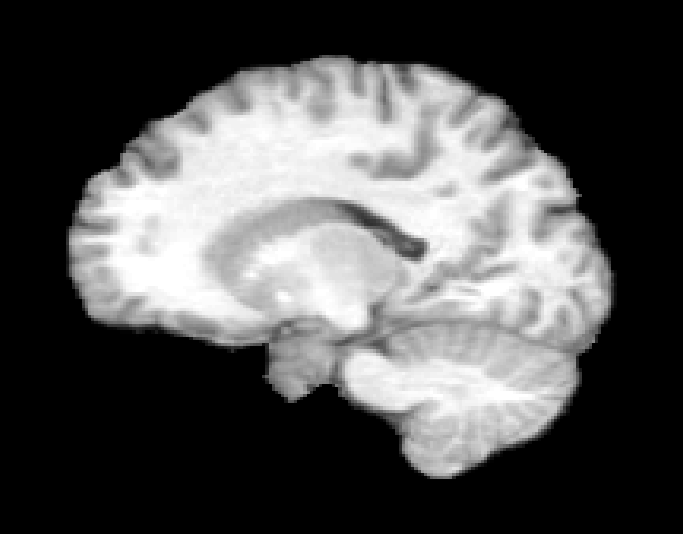} &
		\includegraphics[width=\linewidth]{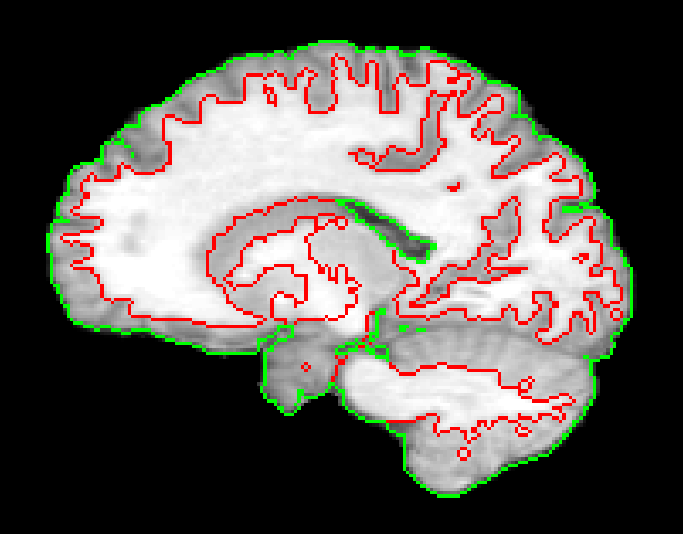} &
		\includegraphics[width=\linewidth]{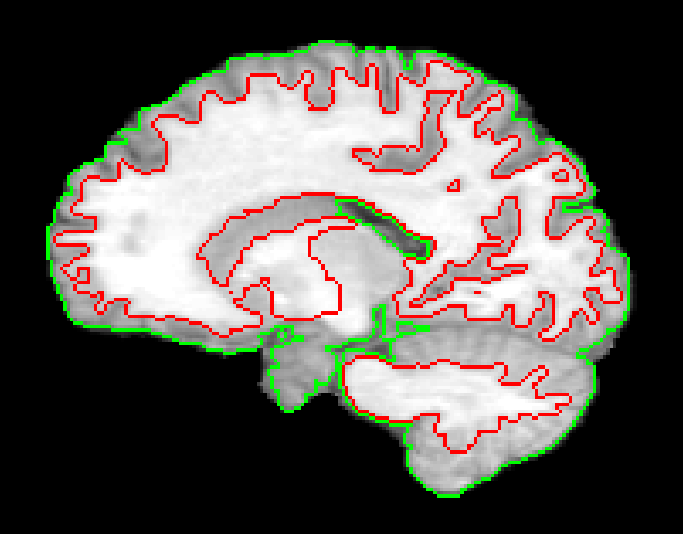} \\
		\rotatebox{90}{coronal view} &
		\includegraphics[width=\linewidth]{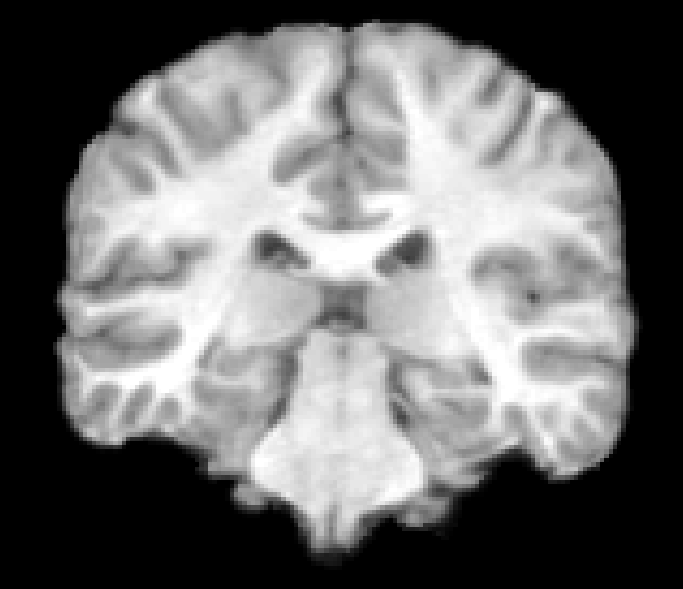} &
		\includegraphics[width=\linewidth]{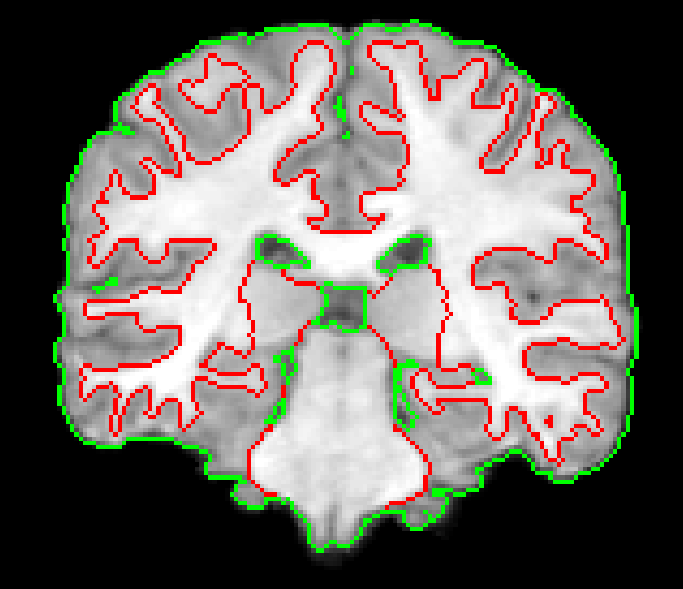} &
		\includegraphics[width=\linewidth]{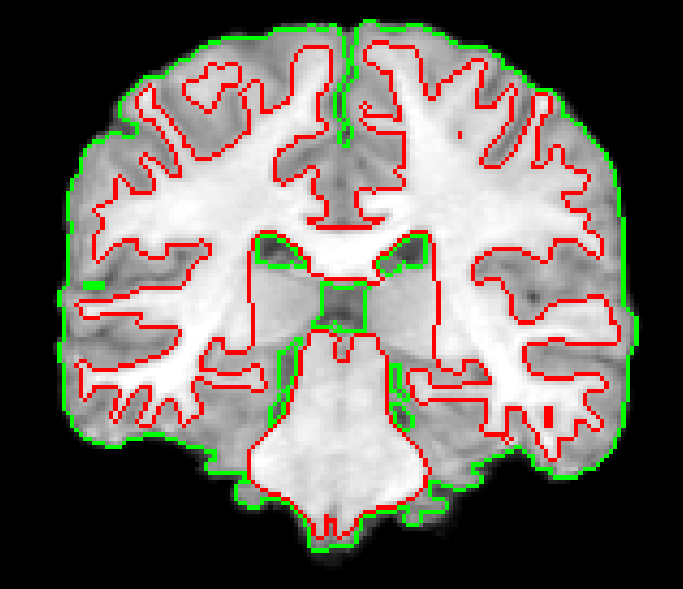} \\
		\rotatebox{90}{axial view} &
		\includegraphics[width=\linewidth]{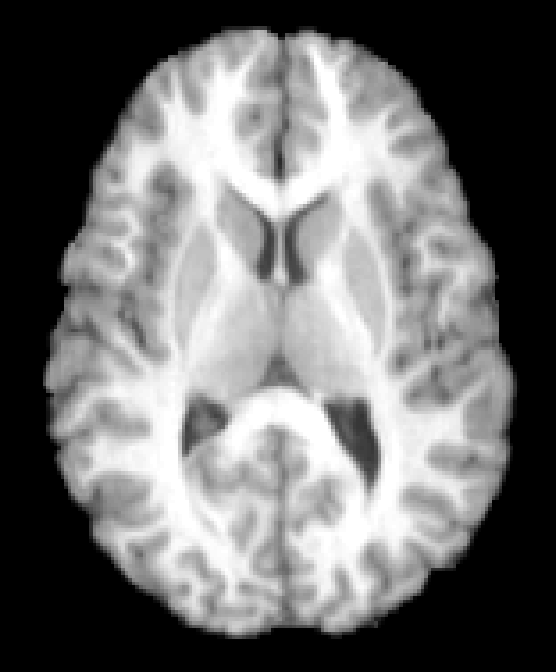} &
		\includegraphics[width=\linewidth]{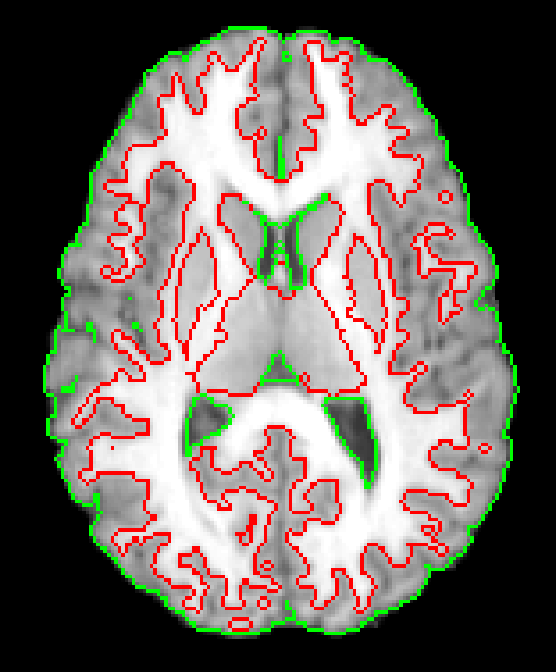} &
		\includegraphics[width=\linewidth]{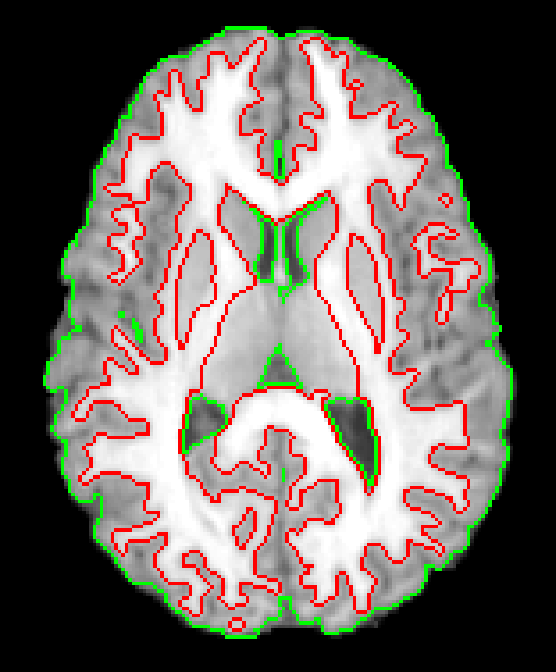} \\
	\end{tabular}
	\caption{Example segmentation using proposed GVF-based shape prior. The red contour is WM-GM boundary. The green contour is GM-CSF boundary.}
	\label{fig:brainContourResult}
\end{figure*}

\begin{figure*}
	\centering 
	\newcommand{\mysize}{0.28}
	\subfloat[\xxx{original input}]{
		\includegraphics[width=\mysize\linewidth]{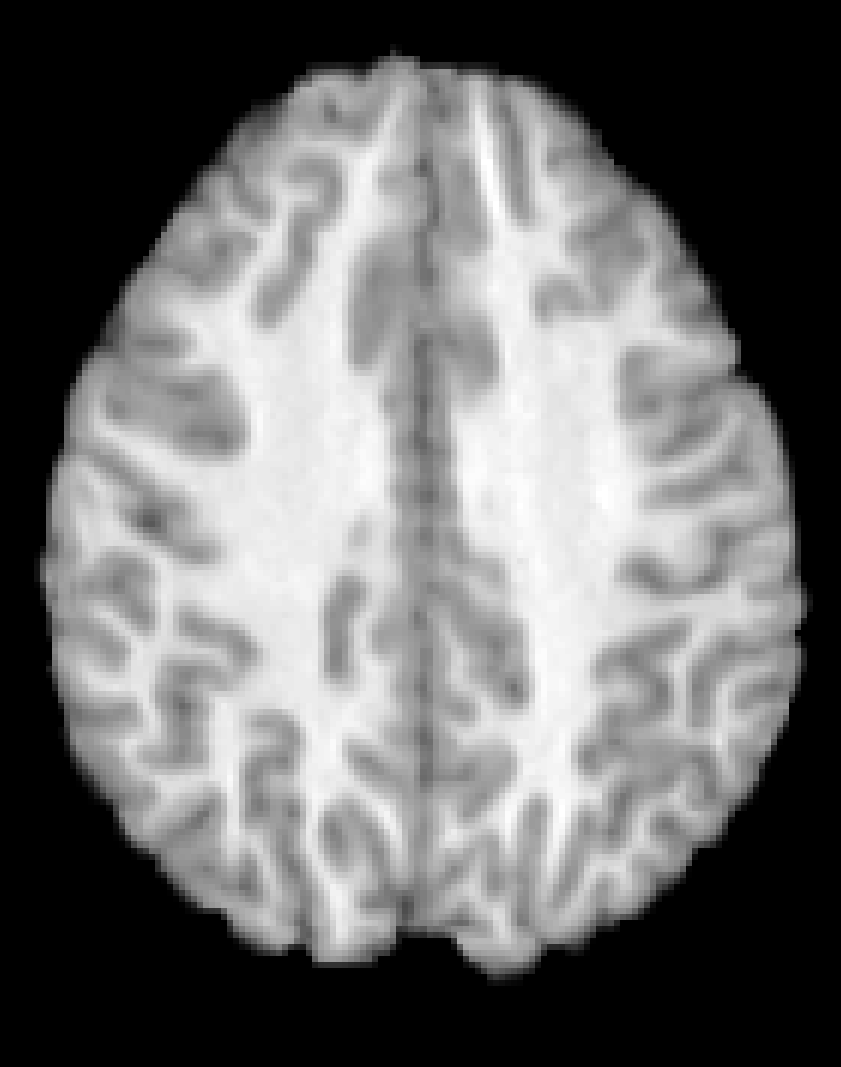} 
		\label{subfig:brainCostInput}}
	\hfill
	\subfloat[\xxx{region term for WM}]{
		\includegraphics[width=\mysize\linewidth]{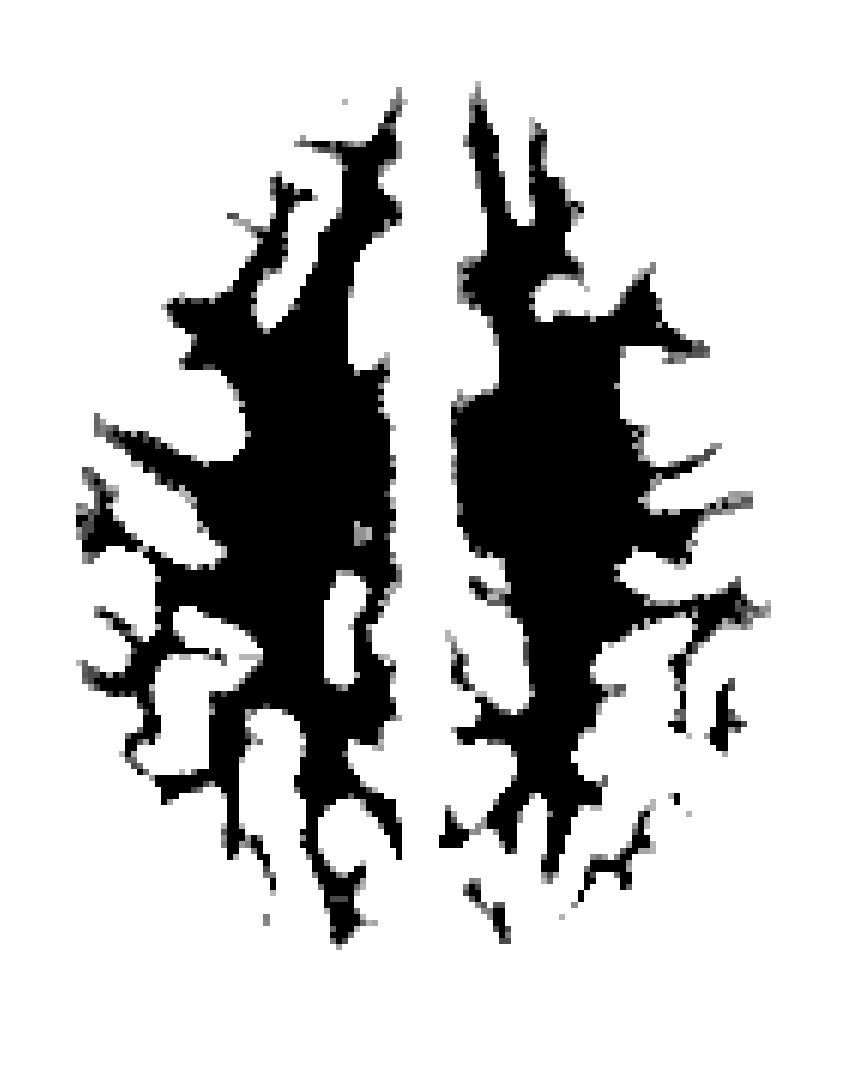} 
		\label{subfig:brainCostRegionTerm}}
	\hfill
	\subfloat[\xxx{GVF shape prior}]{
		\includegraphics[width=\mysize\linewidth]{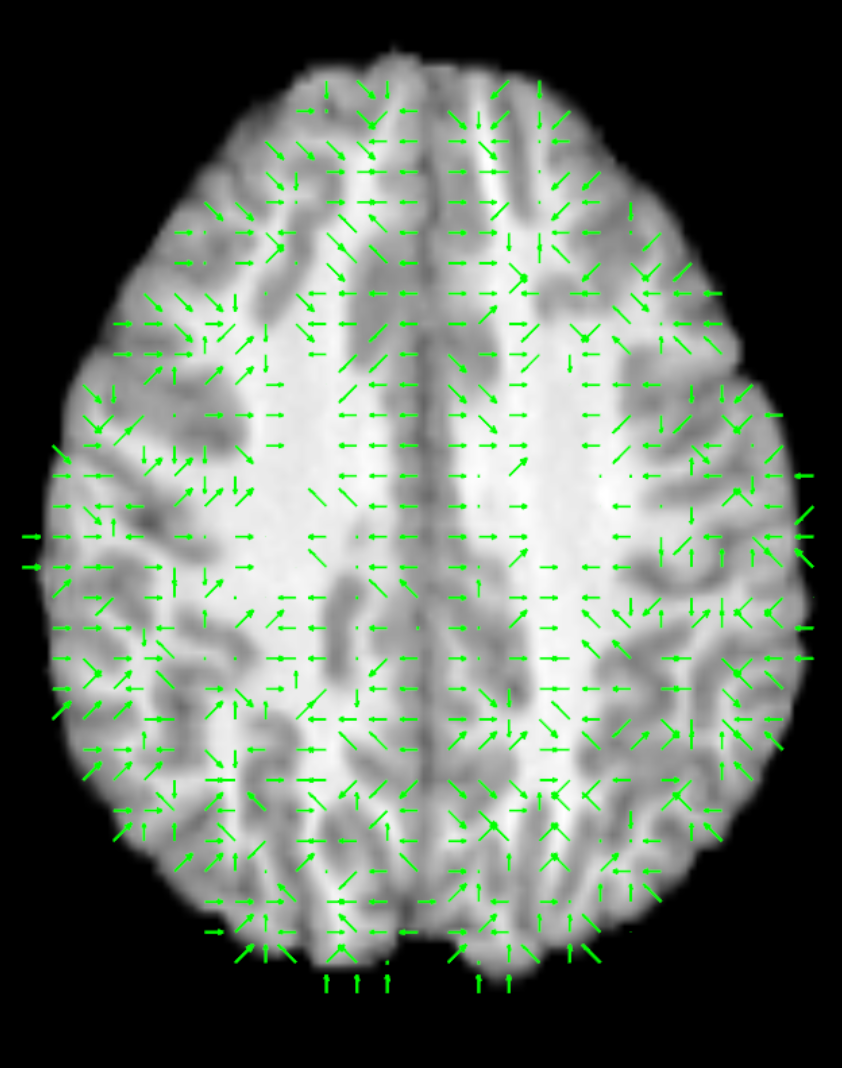} 
		\label{subfig:brainCostGVF}}	\\[-2ex]
	\subfloat[\xxx{sigmoid transformed to enhance WM-GM contrast}]{
		\includegraphics[width=\mysize\linewidth]{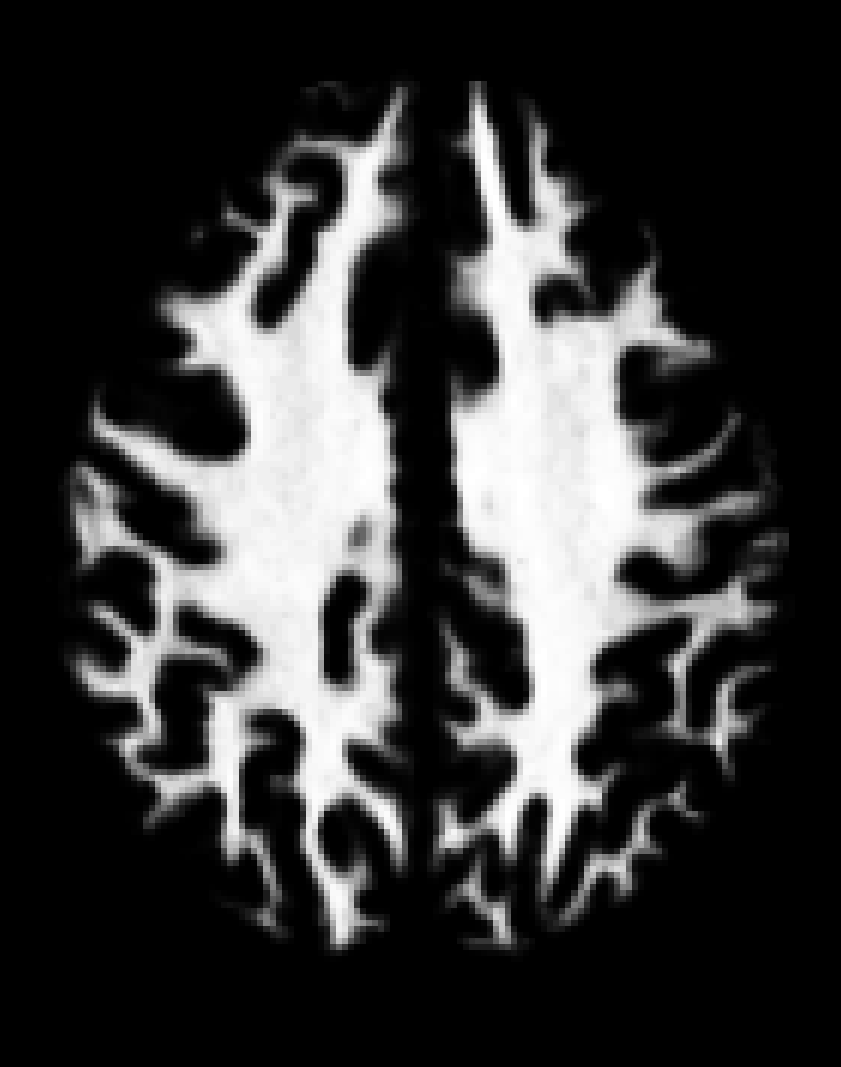} 
		\label{subfig:brainCostSigmoid}}
	\hfill
	\subfloat[\xxx{boundary term for verticle edges}]{
		\includegraphics[width=\mysize\linewidth]{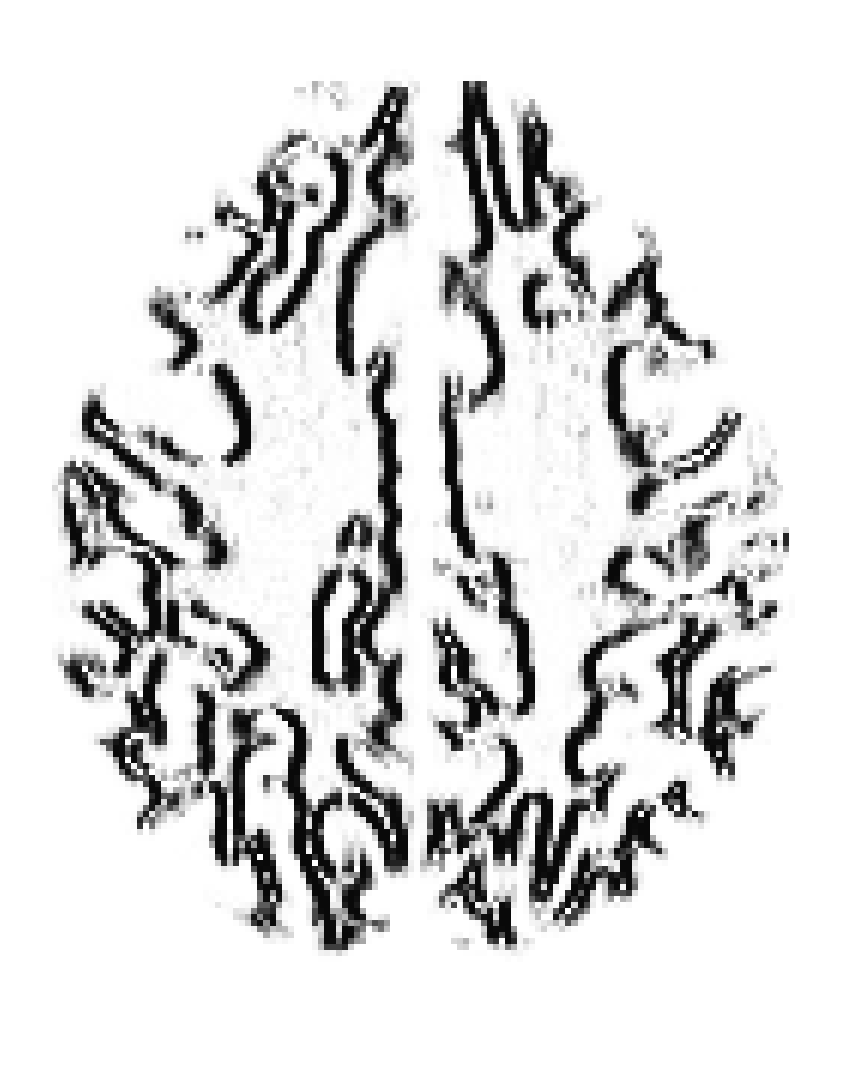} 
		\label{subfig:brainCostBndTermVertical}}
	\hfill
	\subfloat[\xxx{boundary term for horizontal edges}]{
		\includegraphics[width=\mysize\linewidth]{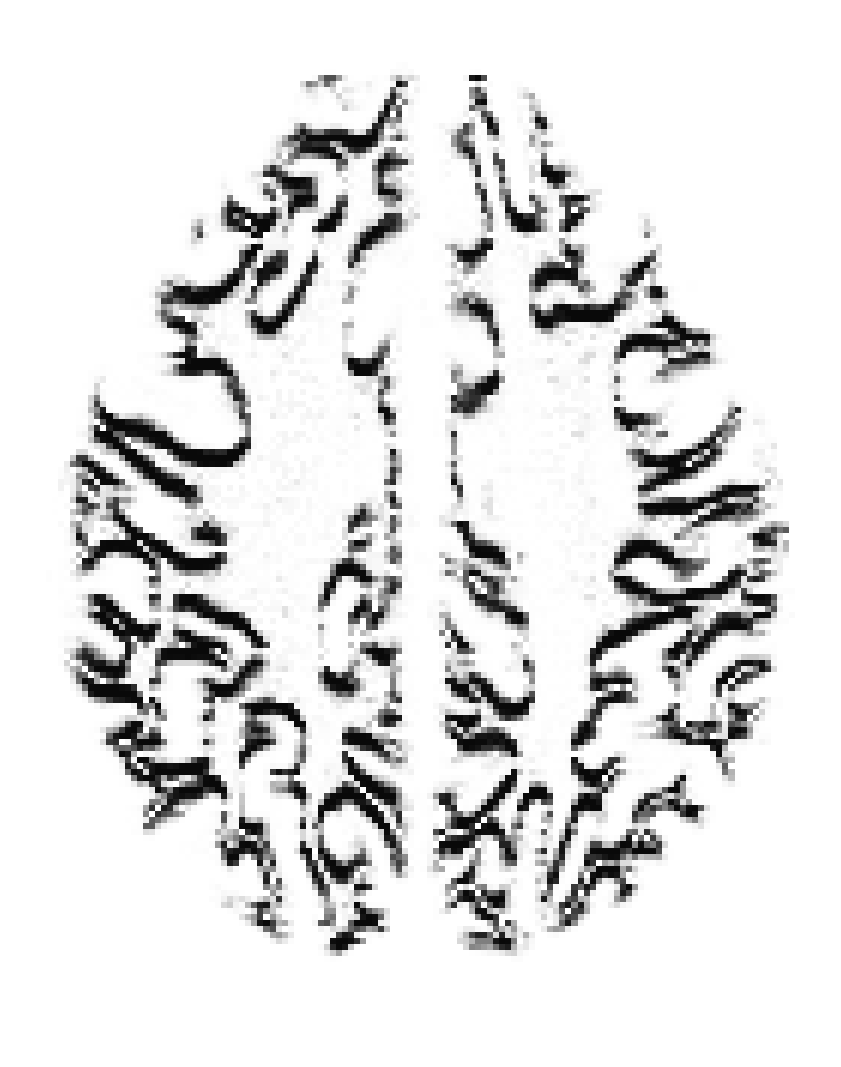} 
		\label{subfig:brainCostBndTermHorizontal}}	\\[-2ex]
	\subfloat[\xxx{pre-segmentation}]{
		\includegraphics[width=\mysize\linewidth]{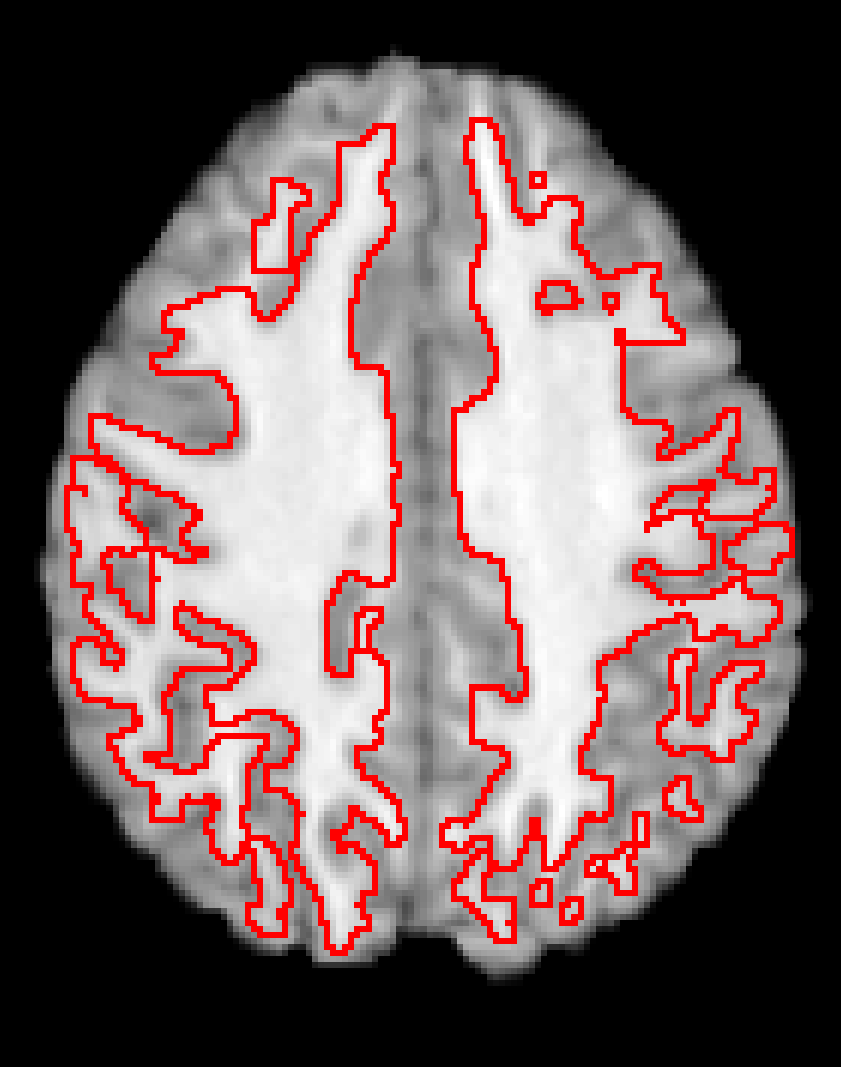} 
		\label{subfig:brainCostInputPreseg}}
	\hfill
	\subfloat[\xxx{manual contour}]{
		\includegraphics[width=\mysize\linewidth]{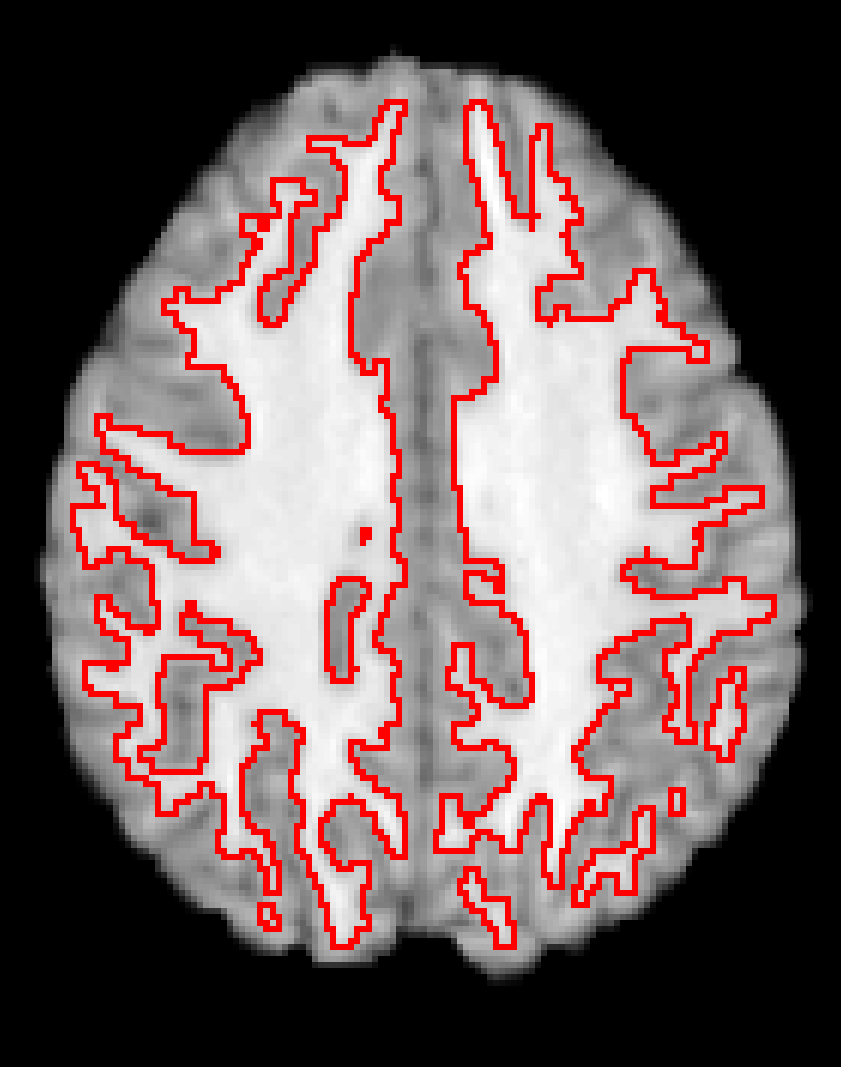} 
		\label{subfig:brainCostInputGT}}
	\hfill
	\subfloat[\xxx{proposed}]{
		\includegraphics[width=\mysize\linewidth]{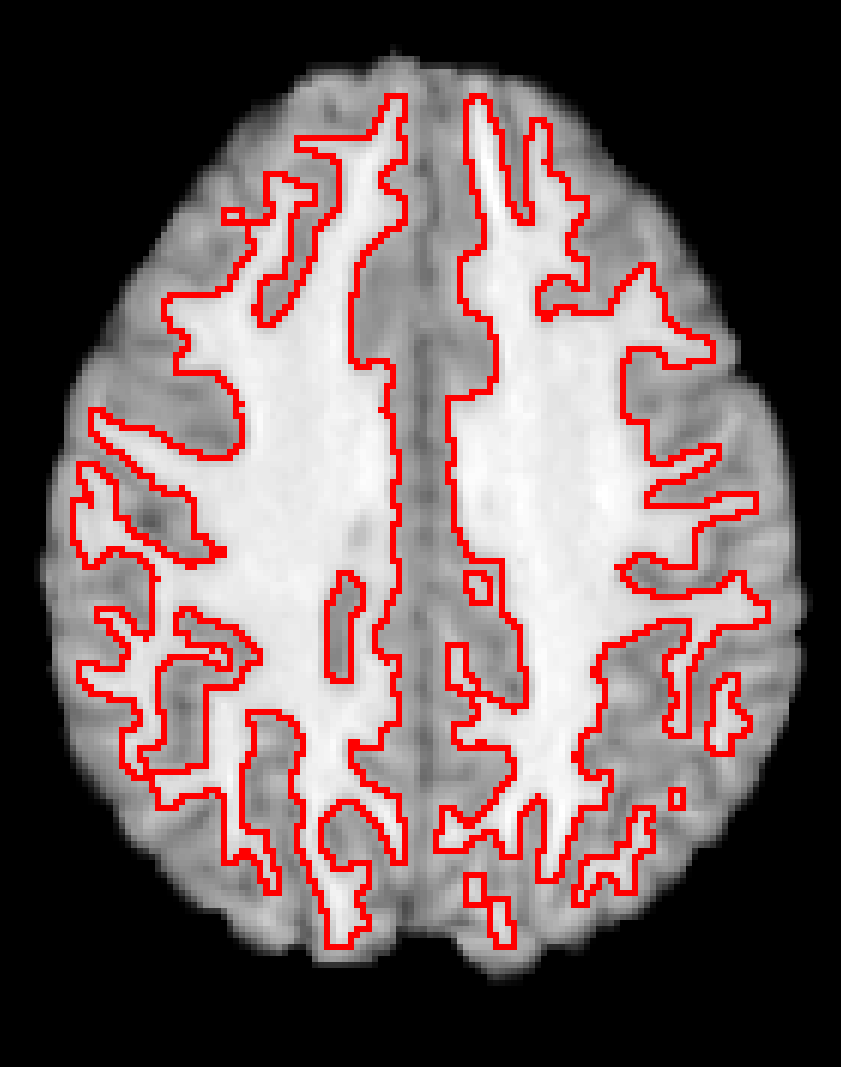} 
		\label{subfig:brainCostInputProposed}}	\\[-2ex]
	\subfloat[\xxx{pre-segmentation}]{
		\includegraphics[width=\mysize\linewidth]{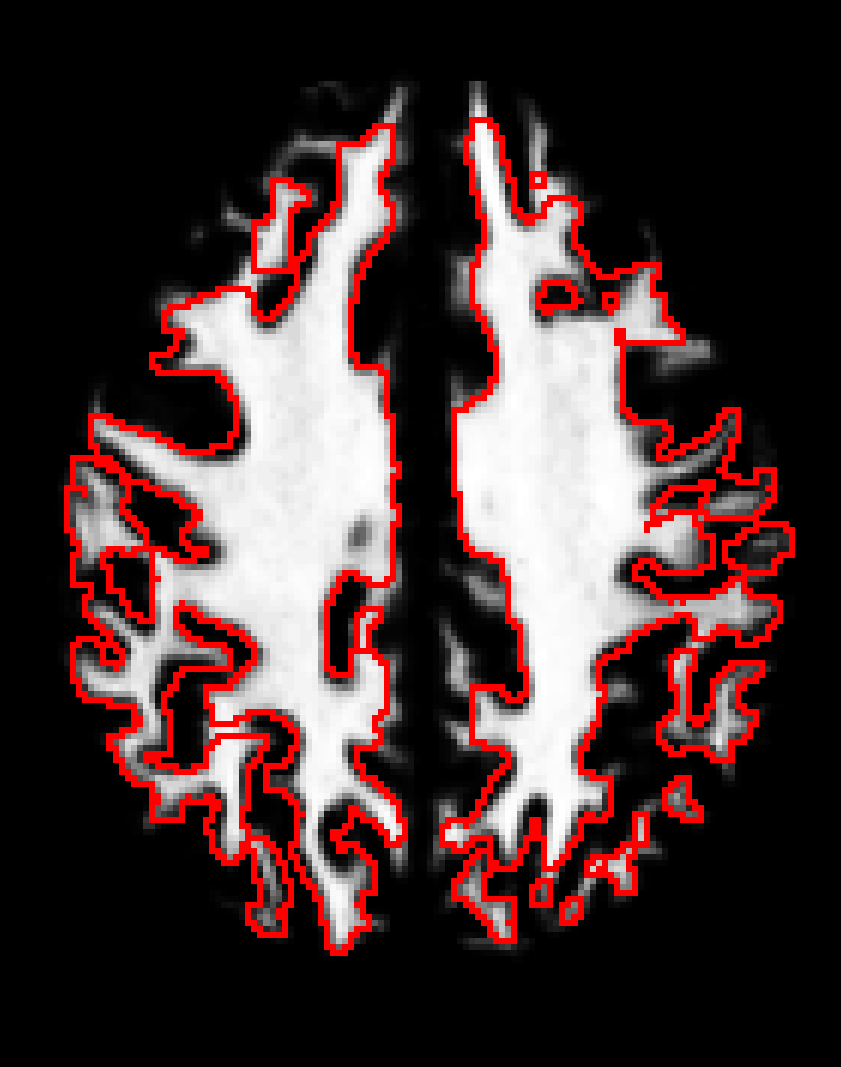} 
		\label{subfig:brainCostSigmoidPreseg}}
	\hfill
	\subfloat[\xxx{manual contour}]{
		\includegraphics[width=\mysize\linewidth]{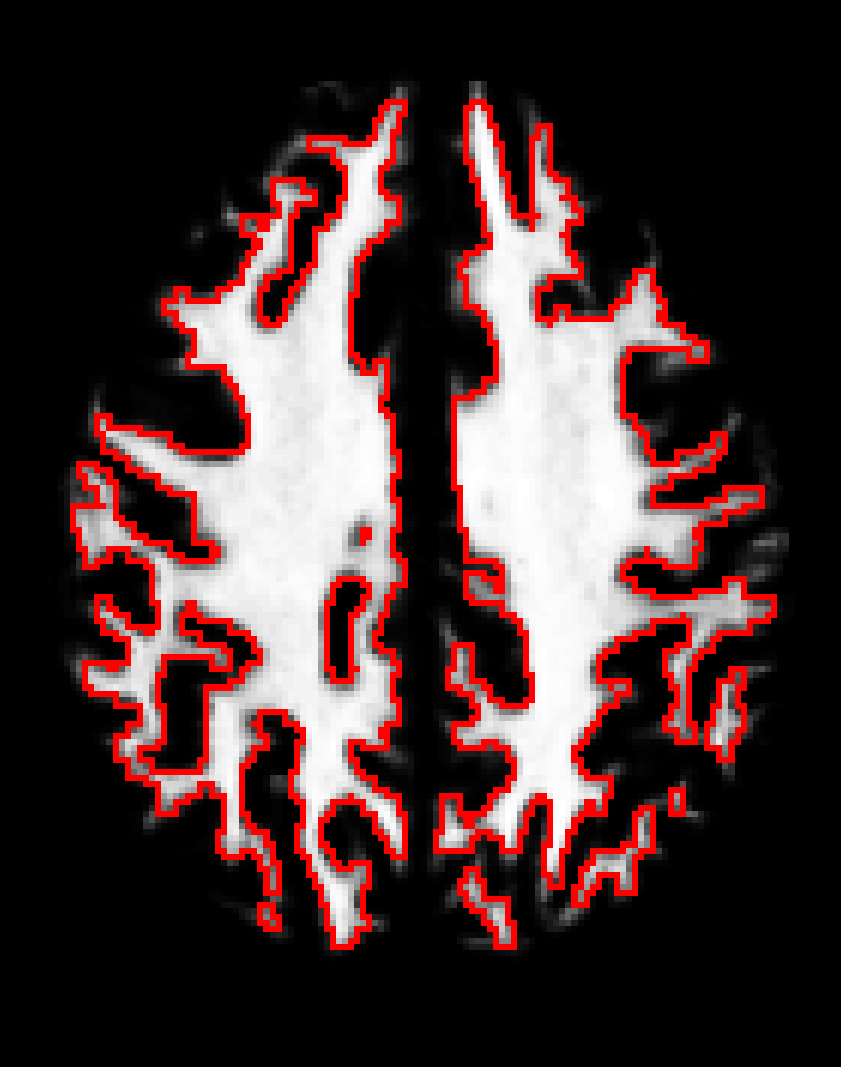} 
		\label{subfig:brainCostSigmoidGT}}
	\hfill
	\subfloat[\xxx{proposed}]{
		\includegraphics[width=\mysize\linewidth]{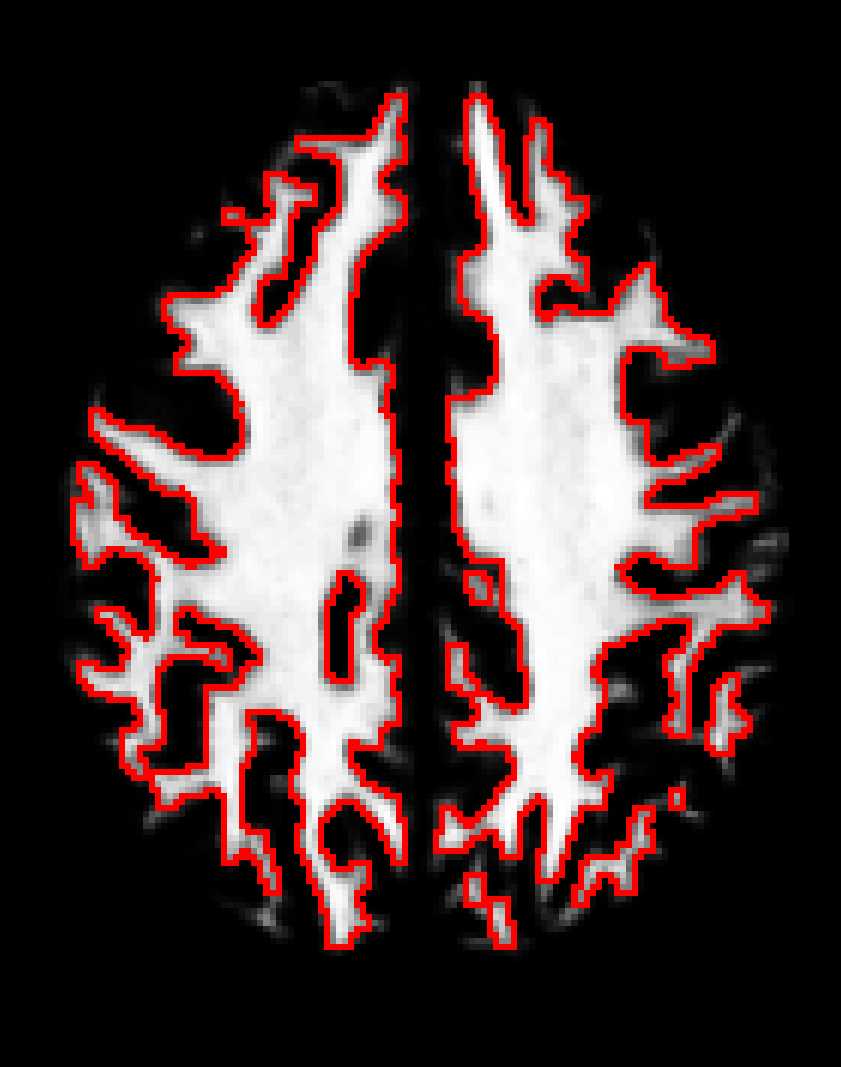} 
		\label{subfig:brainCostSigmoidProposed}}
	\caption{\xxx{Example cost images and GVF shape priors for segmenting WM (Fig.\protect\subref{subfig:brainCostRegionTerm}\protect\subref{subfig:brainCostGVF}\protect\subref{subfig:brainCostBndTermVertical}\protect\subref{subfig:brainCostBndTermHorizontal}), and the corresponding pre-segmentation (Fig.\protect\subref{subfig:brainCostInputPreseg}\protect\subref{subfig:brainCostSigmoidPreseg}), manual contour (Fig.\protect\subref{subfig:brainCostInputGT}\protect\subref{subfig:brainCostSigmoidGT}), and final segmentation by proposed method (Fig.\protect\subref{subfig:brainCostInputProposed}\protect\subref{subfig:brainCostSigmoidProposed}), overlaid on original input image (Fig.\protect\subref{subfig:brainCostInput}) and the sigmoid transformed image to highlight the edge between WM and GM (Fig.\protect\subref{subfig:brainCostSigmoid}). To improve visibility in \protect\subref{subfig:brainCostGVF}, GVF shape prior at one voxel is shown within every 5-by-5 voxel neighborhood. }}
	\label{fig:brainCost}
\end{figure*}

One example segmentation from different views is shown in Fig.\ref{fig:brainContourResult}. 
The three columns show the original image, the manual contour and the segmentation of our 
GVF-based shape prior method, respectively. 
We can see the segmentation aligns well with the manual contour. 
\xxx{In Fig.\ref{subfig:brainCostInputProposed}, \ref{subfig:brainCostSigmoidProposed}, we also show the proposed method's segmentation of WM computed from the region term, boundary terms, and GVF shape priors shown in Fig.\ref{subfig:brainCostRegionTerm}-\ref{subfig:brainCostBndTermHorizontal}. Compared to the pre-segmentation (Fi.g\ref{subfig:brainCostInputPreseg},\ref{subfig:brainCostSigmoidPreseg}), the proposed method's result is aligning better with the reference manual contour (Fig.\ref{subfig:brainCostInputGT}, \ref{subfig:brainCostSigmoidGT}). }

\begin{figure}
	\centering
	\includegraphics[width=\linewidth]{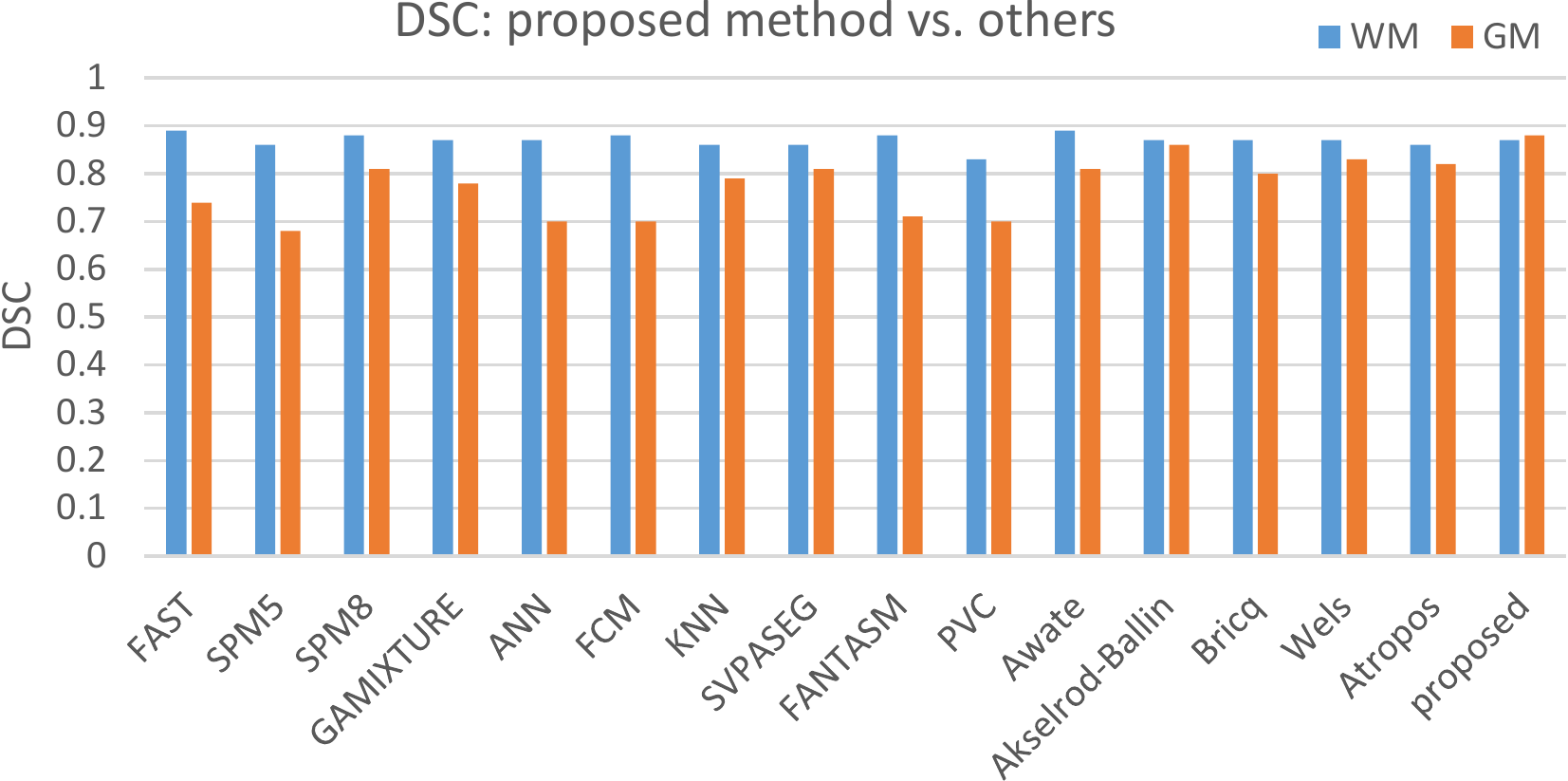}
	\caption{Brain tissue segmentation DSC of proposed method compared to other state-of-art methods, generated from statistics reported in Table.\ref{tb:dscOtherMethods}. The proposed method achieves accurate segmentation on both types of tissue, while most other methods are performing well on WM, but relatively poor on GM.}
	\label{fig:dscOtherMethods}
\end{figure}

Quantitatively we compare the Dice coefficient (DSC) of WM and GM regions with various published methods (Sec.\ref{subsec:brainDataCompareMethods}). Table.\ref{tb:dscOtherMethods} lists the DSC means and standard deviations of various state-of-art methods and the proposed method. Our method is giving the best accuracy on GM, with a large margin compared to most of the other methods. The WM accuracy is also very competitive. In fact, if we visualize the results in a column chart (Fig.\ref{fig:dscOtherMethods}), it is obvious that the proposed method gives accurate segmentation of both WM and GM. In contrast, most other methods performs well on WM, but relatively poorly on GM. 
The surface distance error metric (ASSD) is not reported since many published methods did not report them. 

\xxx{For each brain MRI scan, it takes 1 hour 31 minutes for BRAINSABC to do the pre-segmentation. Given the pre-segmentation, it only takes, on average, 1 minutes 40 seconds to compute the GVF-based shape prior, the energy terms, and the final segmentation. In total, it takes an average of 1 hour 32 minutes and 40 seconds to process one scan. }


\begin{table}
	\centering 
	\caption{Proposed method's DSC compared to other state-of-art methods on WM and GM in brain tissue segmentation. Bold items are the best accuracy for each tissue type among all methods. Our method returns the most accurate GM, and is very competitive on WM accuracy.}
	\label{tb:dscOtherMethods}
	\begin{tabular}{c|cc}
		\hline
		method & WM & GM \\
		\hline
		FAST~\cite{Zhang2001} & 0.89 $\pm$ 0.02  & 0.74 $\pm$ 0.04\\
		SPM5~\cite{Ashburner2005} & 0.86  $\pm$ 0.02 & 0.68 $\pm$ 0.07 \\
		SPM8~\cite{ashburner2012spm8} & 0.88  $\pm$ 0.01 & 0.81 $\pm$ 0.02 \\
		GAMIXTURE~\cite{Tohka2007} & 0.87 $\pm$ 0.02 & 0.78 $\pm$ 0.08 \\
		ANN~\cite{Tian2007} & 0.87 $\pm$ 0.03 & 0.70 $\pm$ 0.07 \\
		FCM~\cite{Pham2001} & 0.88 $\pm$ 0.03 & 0.70 $\pm$ 0.06 \\
		KNN~\cite{DeBoer2009} & 0.86 $\pm$ 0.03 & 0.79 $\pm$ 0.03 \\
		SVPASEG~\cite{Tohka2010} & 0.86 $\pm$ 0.02 & 0.81 $\pm$ 0.03 \\
		FANTASM~\cite{Pham} & 0.88 $\pm$ 0.03 & 0.71 $\pm$ 0.06 \\
		PVC~\cite{Shattuck2001} & 0.83 $\pm$ 0.07 & 0.70 $\pm$ 0.08 \\
		Awate \textit{et al.}
		~\cite{Awate2006} & \textbf{0.89 $\pm$ 0.02} & 0.81 $\pm$ 0.04 \\
		Akselrod-Ballin \textit{et al.}
		~\cite{Akselrod-Ballin2007} & 0.87 & 0.86 \\
		Bricq \textit{et al.}
		\cite{Bricq2008} & 0.87  $\pm$ 0.02 & 0.80 $\pm$ 0.06 \\
		Wels \textit{et al.}
		~\cite{Wels2011} & 0.87  $\pm$ 0.05 & 0.83 $\pm$ 0.12 \\
		\xxx{Atropos~\cite{Avants2011}} & \xxx{0.86 $\pm$ 0.03} & \xxx{ 0.82 $\pm$ 0.03 }\\
		\hline
		proposed & 0.87 $\pm$ 0.03 & \textbf{0.88 $\pm$ 0.02 }\\
		\hline 
	\end{tabular}
\end{table}

\subsection{Brain Segmentation: Sensitivity to Pre-segmentation}
\label{sec-brain-sensitivity}
Since the shape prior is defined based on a pre-segmentation, it is natural to see how sensitive the method is with respect to the pre-segmentation. To study the sensitivity, we artificially warped the image on a $20\times 20 \times 20$ grid using B-Spline. At each B-Spline grid point, a random Gaussian noise deformation force with a mean of zero and a standard deviation of $\sigma^{ptb}$ along each dimension is applied. Then the proposed GVF-based shape prior segmentation algorithm is run using a pre-segmentation perturbed with various perturbation magnitudes $\sigma^{ptb}$.
The DSC and ASSD between the perturbed pre-segmentation and the manual expert contour are measured to quantify the deformation magnitude (reported as `pre-seg DSC/ASSD' in Table.\ref{tb:brainSensitivity}). The DSC and ASSD between the output of the proposed algorithm and the manual expert contours are reported to validate how sensitive the proposed method is to the pre-segmentation (reported as `final DSC/ASSD' in Table.\ref{tb:brainSensitivity}). 

\begin{figure*}
	\centering 
	\subfloat[DSC of perturbed pre-seg vs. manual contour]{
		\includegraphics[width=.47\linewidth]{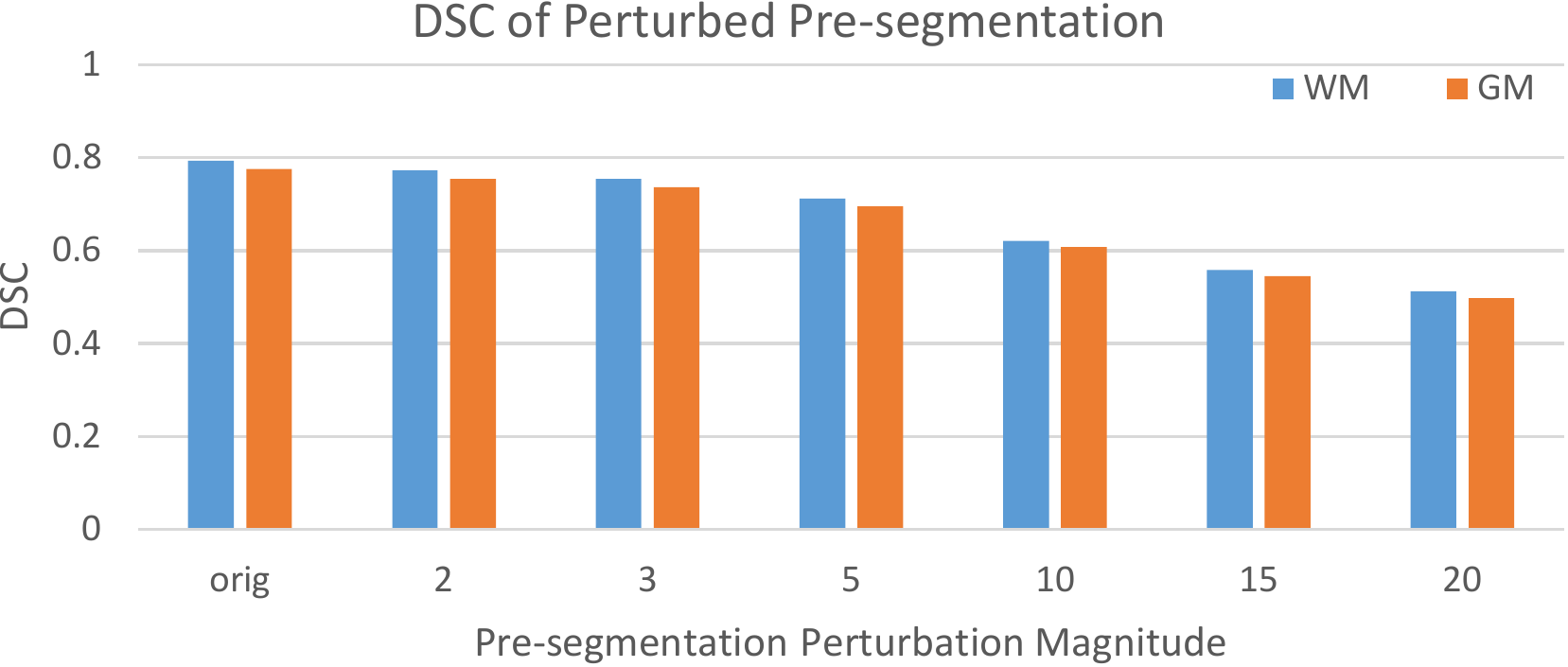} 
		\label{subfig:perturbPresegDSC}}
	\hfill
	\subfloat[ASSD of perturbed pre-seg vs. manual contour]{
		\includegraphics[width=.47\linewidth]{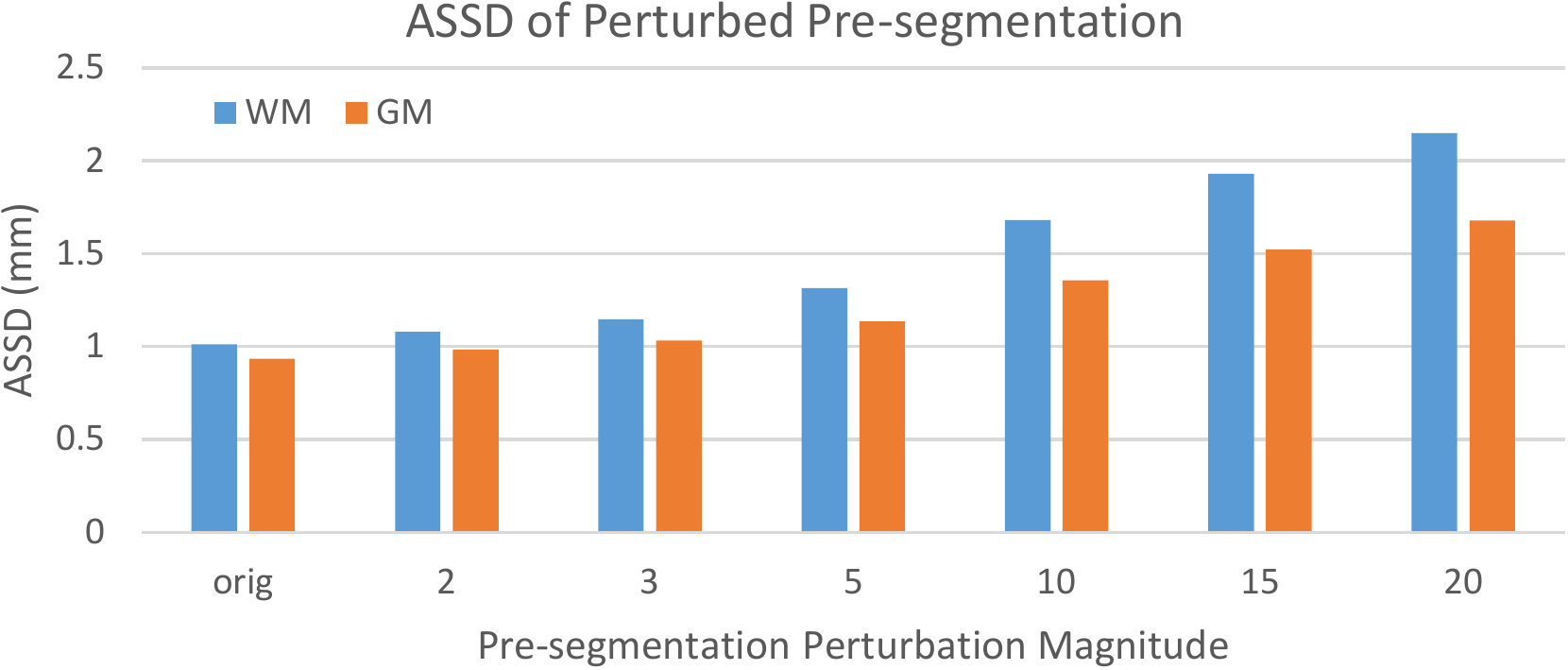} 
		\label{subfig:perturbPresegASSD}}\\
	\subfloat[DSC of proposed method using perturbed pre-seg]{
		\includegraphics[width=.47\linewidth]{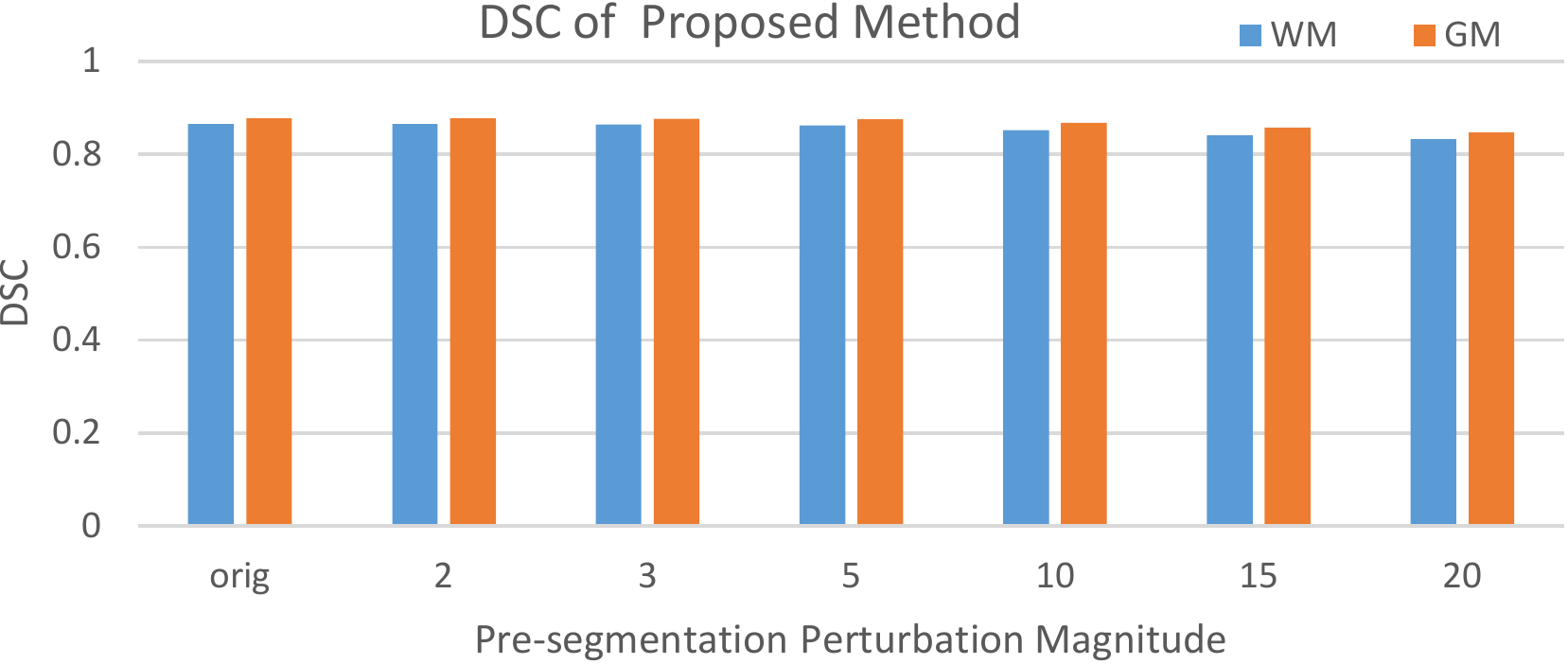} 
		\label{subfig:segWithPerturbDSC}}
	\hfill
	\subfloat[ASSD of proposed method using pertrubed pre-seg]{
		\includegraphics[width=.47\linewidth]{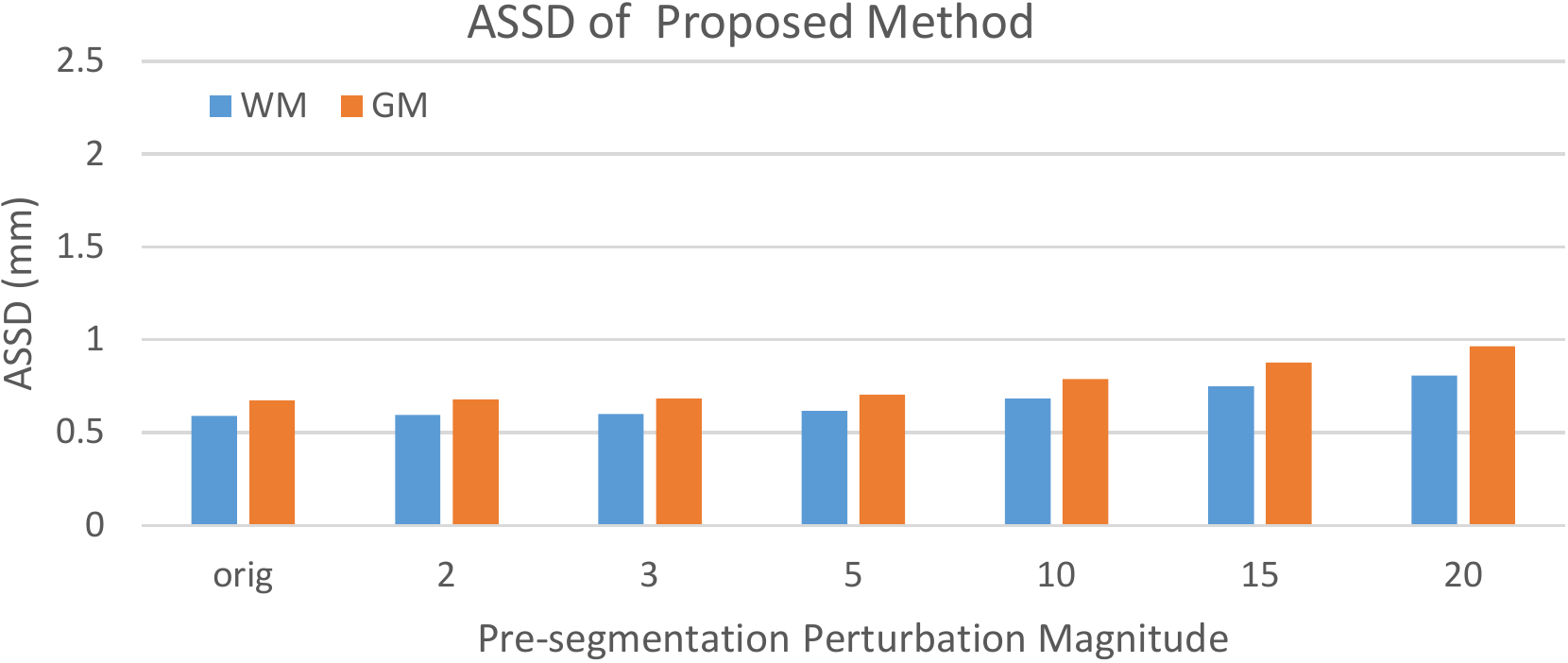} 
		\label{subfig:segWithPerturbASSD}}
	\caption{Sensitivity of the proposed method with respect to pre-segmentation deformation (generated from Table.\ref{tb:brainSensitivity}.). As the deformation greatly deviates from the pre-segmentation, the proposed method only suffers a mild decrease in performance, for both tissue types and both metrics.}
	\label{fig:brainSensitivity}
\end{figure*}

As the perturbation magnitude parameter $\sigma^{ptb}$ increases from 0 to 5, the output segmentation of the proposed method  barely change at all, even though the DSC between the deformed pre-segmentation and manual contour dropped by about 0.04 for both WM and GM. As the perturbation magnitude increases even larger, the accuracy of the proposed method starts to decrease more significantly. But even at the largest perturbation magnitude ($\sigma^{ptb}=20$), the DSC only drops around 0.03 for both WM and GM, while the deformed pre-segmentation only has 0.512 and 0.498 in DSC. In terms of surface distance error ASSD, while the pre-segmentation is even perturbed by 1.14 $mm$ on WM and 0.75 $mm$ on GM,  the output ASSD's only drop by 0.22 and 0.30 $mm$, respectively. 
Fig.\ref{fig:brainSensitivity} visually conveys this trend. In the first row of Fig.\ref{fig:brainSensitivity}, as the deformed pre-segmentation rapidly deviates from the original pre-segmentation (Fig.\ref{subfig:perturbPresegDSC},\ref{subfig:perturbPresegASSD}), the  segmentation accuracy of the proposed method only decreases mildly (Fig.\ref{subfig:segWithPerturbDSC},\ref{subfig:segWithPerturbASSD}).  

\begin{table}[]
	\centering
	\caption{Sensitivity of proposed method with respect to deformation in pre-segmentation. The first column is the perturbation magnitude parameter $\sigma^{ptb}$ value. The surface error (ASSD) values are reported in unit $mm$.}
	\label{tb:brainSensitivity}
	\begin{tabular}{c||cc|cc||cc|cc}
		\hline
		\multirow{2}{*}{\parbox{0.8cm}{\centering ptb mag}} & \multicolumn{2}{c|}{pre-seg DSC} & \multicolumn{2}{c||}{final DSC} & \multicolumn{2}{c|}{pre-seg ASSD} & \multicolumn{2}{c}{final ASSD} \\
		& WM              & GM             & WM              & GM              & WM              & GM              & WM               & GM              \\ \hhline{=#====#====}
		0 & 0.794 & 0.775 & 0.866 & 0.878 & 1.01 & 0.93 & 0.59 & 0.67 \\ \hline
		2    & 0.772 & 0.754 & 0.865 & 0.877 & 1.08 & 0.99 & 0.59 & 0.68 \\ \hline
		3    & 0.754 & 0.736 & 0.864 & 0.877 & 1.15 & 1.03 & 0.60 & 0.68 \\ \hline
		5    & 0.712 & 0.696 & 0.862 & 0.876 & 1.32 & 1.13 & 0.62 & 0.71 \\ \hline
		10   & 0.621 & 0.608 & 0.852 & 0.868 & 1.68 & 1.36 & 0.69 & 0.79 \\ \hline
		15   & 0.559 & 0.545 & 0.841 & 0.857 & 1.93 & 1.52 & 0.75 & 0.88 \\ \hline
		20   & 0.512 & 0.498 & 0.832 & 0.848 & 2.15 & 1.68 & 0.81 & 0.97 \\ \hline
	\end{tabular}
\end{table}

\begin{figure*}
	\centering 
	\subfloat[original]{
		\includegraphics[width=.22\linewidth]{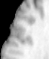} 
		\label{subfig:brainSliceOrig}}
	\hfill
	\subfloat[manual WM contour]{
		\includegraphics[width=.22\linewidth]{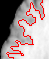} 
		\label{subfig:brainSliceOrigGt}}
	\hfill
	\subfloat[pre-segmentation binary mask]{
		\includegraphics[width=.22\linewidth]{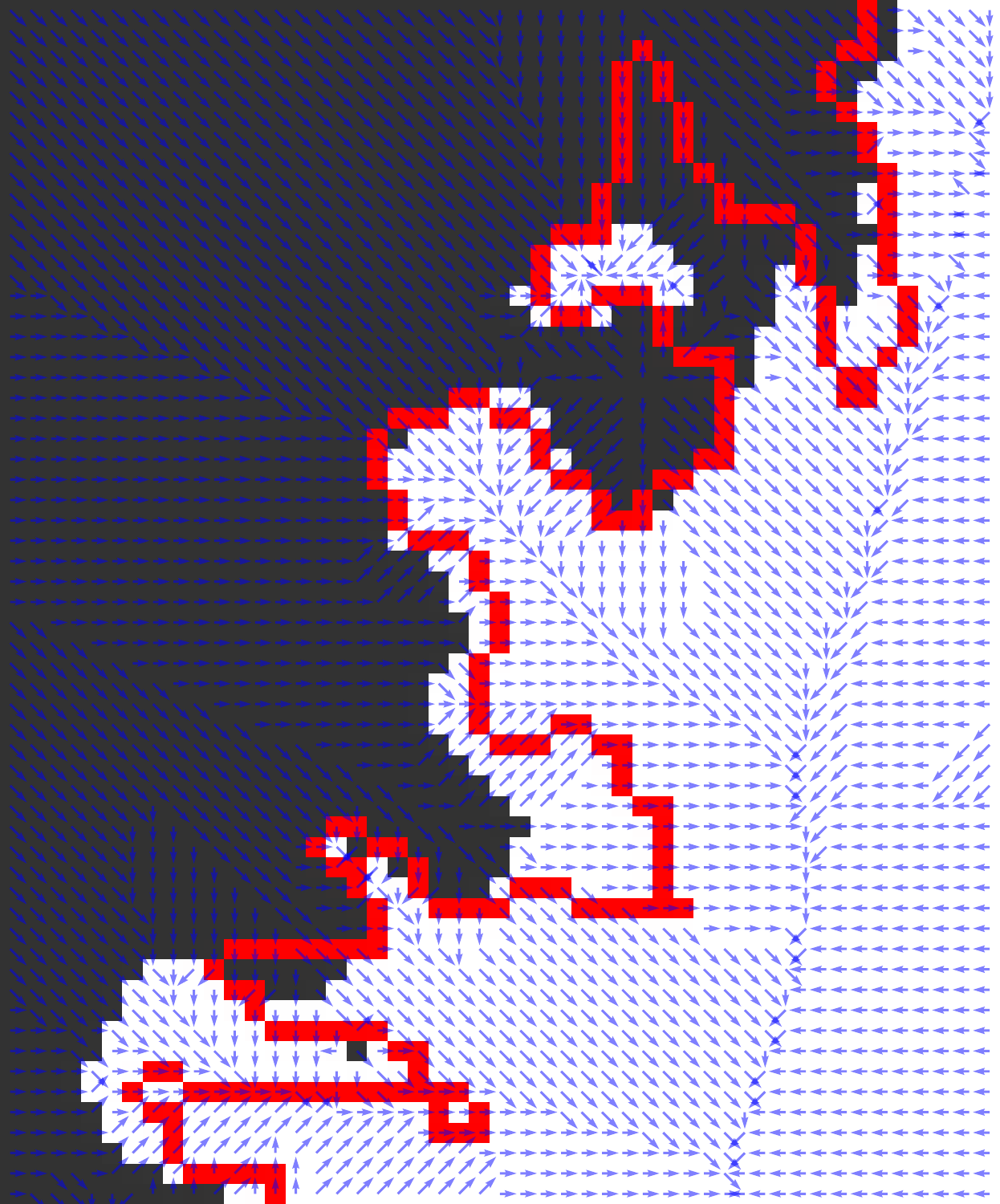} 
		\label{subfig:brainSlicePreseg}}
	\hfill
	\subfloat[seg with GVF-based shape priors]{
		\includegraphics[width=.22\linewidth]{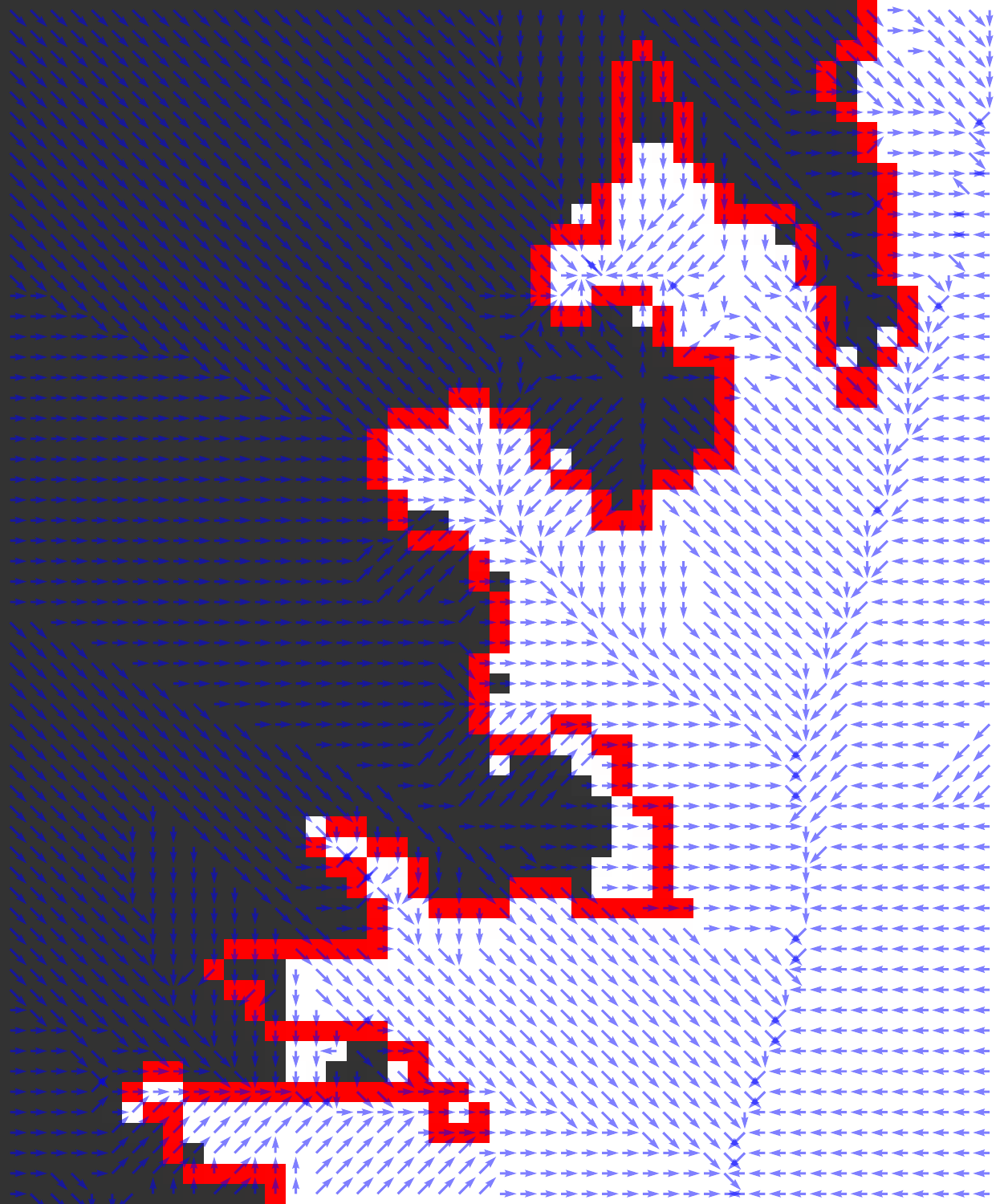} 
		\label{subfig:brainSliceGvfSeg}}
	\caption{The GVF-based shape prior is flexible. Even though the pre-segmentation boundary (white mask in \protect\subref{subfig:brainSlicePreseg}) is not quite accurate around the white matter boundary compared to the manual delineation (red contours), the  segmentation using the GVF-based shape prior (white mask in \protect\subref{subfig:brainSliceGvfSeg}) is still able to take advantage of the shape information, and achieve accurate segmentation with respect to the manual contour. Small blue arrows in \protect\subref{subfig:brainSlicePreseg} and \protect\subref{subfig:brainSliceGvfSeg} are the representation of the GVF-based shape prior computed from the pre-segmentation (zoom in to see more details).}
	\label{fig:brainSliceFlexibleGVF}
\end{figure*}

Fig.\ref{fig:brainSliceFlexibleGVF} demonstrates the flexibility of the proposed GVF-based shape prior. Although the pre-segmentation (the white mask in Fig.\ref{subfig:brainSlicePreseg}) is not very accurate compared to the manual delineation of white matter (the red contour), the segmentation using GVF-based shape prior (the white mask Fig.\ref{subfig:brainSliceGvfSeg}) is still able to take advantage of the shape information, and achieve good boundary alignment with the manual contour. The GVF-based shape prior computed from the pre-segmentation is drawn as small blue arrows in Fig.\ref{subfig:brainSlicePreseg} and Fig.\ref{subfig:brainSliceGvfSeg} (you may zoom in to see them clearly). 


\subsection{Bladder/prostate Segmentation: Data and Compared Methods}
For prostate and bladder segmentation, 21 volumetric CT images from different patients with prostate cancer were used. The image spacing ranges from $0.98\times 0.98 \times 3.00 mm^3$ to $1.60 \times 1.60 \times 3.00 mm^3$. Expert manual contours of bladder and prostate are available for each scan. Eight scans randomly selected are used as the training set. The other 13 scans are used as the testing set. All parameters are tuned based on the training set. All reported results are from the testing set. 

Our proposed method is compared against the mesh-based approach to enforce the shape priors~\cite{Song2010miccai}. It iteratively evolves the bladder and prostate {\em a prior} shape meshes using the LOGISMOS method. In each iteration, the bladder and prostate shape meshes are first evolved independently, and then deformed jointly by defining a ``mutually interacting region'' to ensure that the two meshes do not intersect. A soft shape prior is also incorporated in the prostate mesh to encourage the shape conforming to a trained model. 

\subsection{Bladder/prostate Segmentation: Workflow}
\subsubsection{Pre-segmentation} We use the same pre-segmentation used by Song \textit{et al.}~\cite{Song2010miccai}. The prostate shape is relatively consistent across scans, thus a point distribution model built from training images is aligned to each test scan. The Procrustes analysis method is employed to get the prostate pre-segmentation. Due to the large shape variations in bladder, a geodesic active contour method is used to generate the pre-segmentation for bladder. The prostate and bladder are set to be two exclusive objects with a minimum separation distance of 0 $mm$. 

\subsubsection{Energy term design} Similar to the brain tissue segmentation, a random forest model with a four-dimensional feature vector (normalized x/y/z coordinates and intensity) is trained to generate probability maps of bladder and prostate. Due to the poor soft-tissue contrast in CT scans, a Chan-Vese model~\cite{Chan2001} is also trained to more accurately model the intensity difference between prostate and bladder. The two probability maps generated by the random forest model and the Chan-Vase model are added together with equal weight as the data term ($D_p(\cdot)$ in Eq.\eqref{eq:mrf}). The smoothness term design is the same as that in brain tissue segmentation (Sec.\ref{subsec:brainWorkflow}).

\subsection{Bladder/prostate Segmentation: Results}

The  segmentation accuracy of the proposed method is presented in Table.\ref{tb:prostateSong}. A 2-tailed paired $t$-test shows that our method is  significantly better than Song \textit{et al.}'s method~\cite{Song2010miccai} on bladder ($p<0.05$) with both DSC and ASSD metrics.  For the prostate, both methods are not statistically different, while our method shows a better DSC but slightly worse ASSD values. 


\begin{table}[]
	\centering
	\caption{Bladder/prostate segmentation accuracy compared to mesh-based method~\cite{Song2010miccai}.}
	\label{tb:prostateSong}
	\begin{tabular}{c|c|cc}
		\hline 
		metric                & object   & Mesh based method~\cite{Song2010miccai} & proposed \\ \hhline{=|e=|==} 
		\multirow{2}{*}{DSC}  & bladder  & 0.933 $\pm$ 0.015    & \textbf{0.941 $\pm$ 0.017}   \\
		& prostate & 0.824 $\pm$ 0.036     & \textbf{0.835 $\pm$ 0.015}   \\ \hline 
		\multirow{2}{*}{\parbox{1cm}{\centering ASSD (mm)}} & bladder  & 0.89 $\pm$ 0.26      & \textbf{0.79 $\pm$ 0.28}    \\
		& prostate & \textbf{1.48 $\pm$ 0.47}      & 1.64 $\pm$ 0.28   \\
		\hline 
	\end{tabular}
\end{table}

\begin{figure*}
	\centering 
	\begin{tabular}{>{\centering\arraybackslash}m{.02\linewidth}>{\centering\arraybackslash}m{.20\linewidth}>{\centering\arraybackslash}m{.20\linewidth}>{\centering\arraybackslash}m{.20\linewidth}>{\centering\arraybackslash}m{.20\linewidth}}
		& original image & manual contour & Song \textit{et al.}~\cite{Song2010miccai} & proposed method\\
		\rotatebox{90}{3D view} &
		\includegraphics[width=\linewidth]{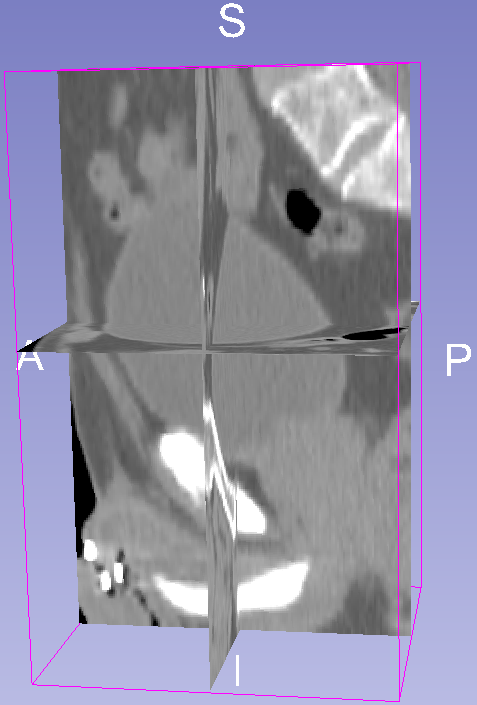} &
		\includegraphics[width=\linewidth]{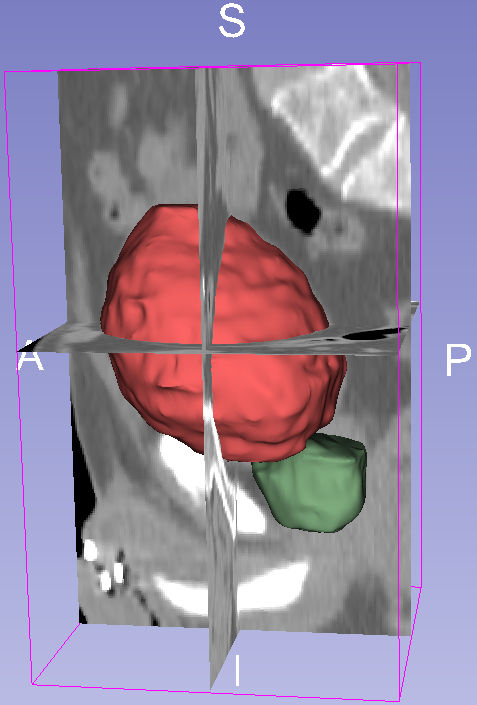} &
		\includegraphics[width=\linewidth]{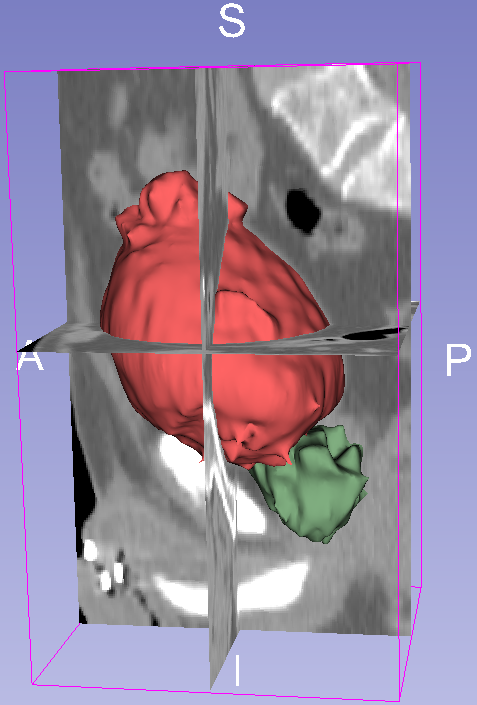} &
		\includegraphics[width=\linewidth]{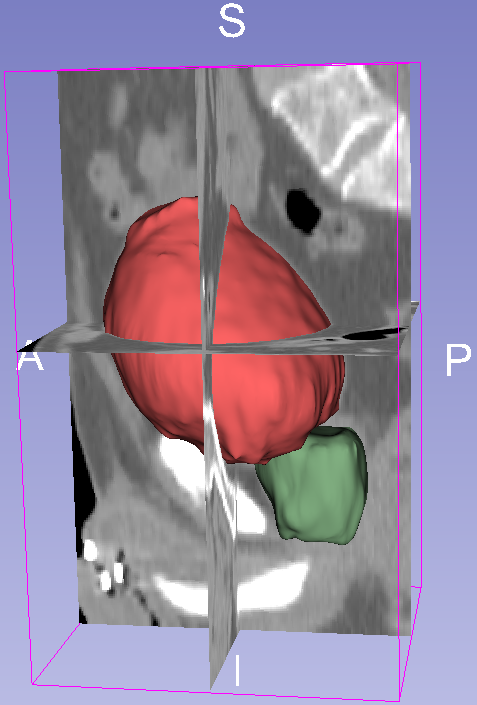} \\
		\rotatebox{90}{sagittal view} &
		\includegraphics[width=\linewidth]{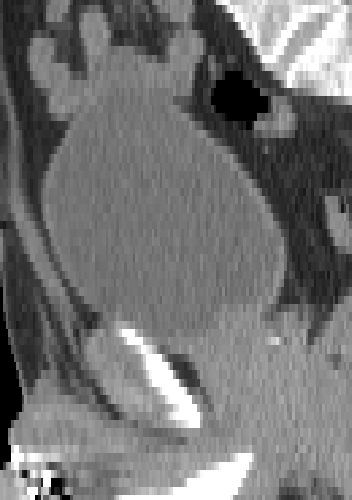} &
		\includegraphics[width=\linewidth]{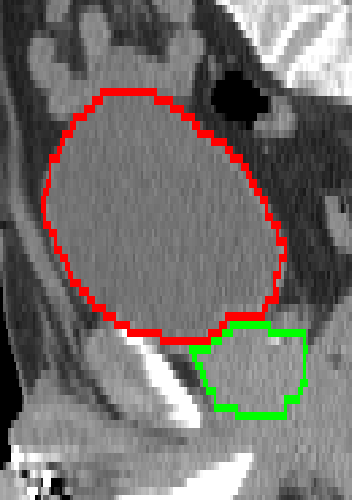} &
		\includegraphics[width=\linewidth]{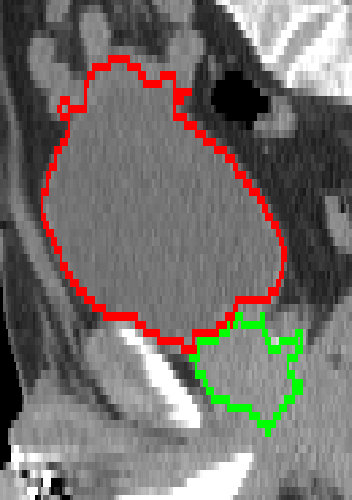} &
		\includegraphics[width=\linewidth]{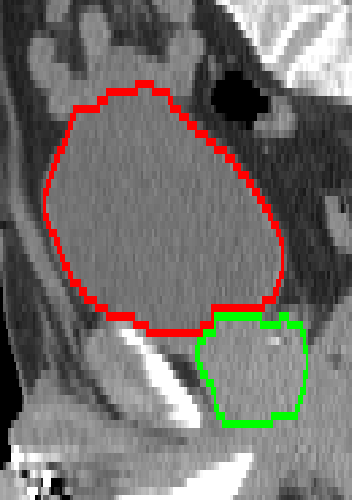} \\
		\rotatebox{90}{coronal view} &
		\includegraphics[width=\linewidth]{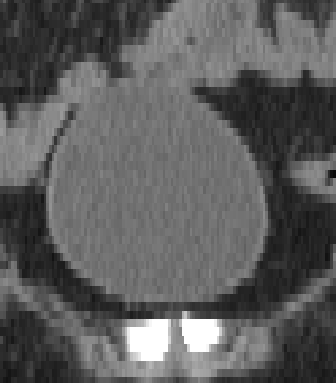} &
		\includegraphics[width=\linewidth]{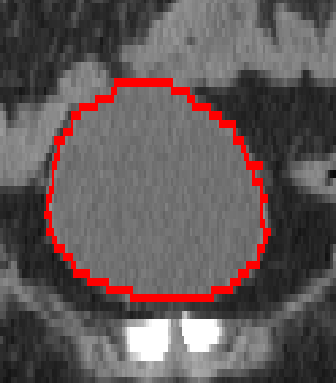} &
		\includegraphics[width=\linewidth]{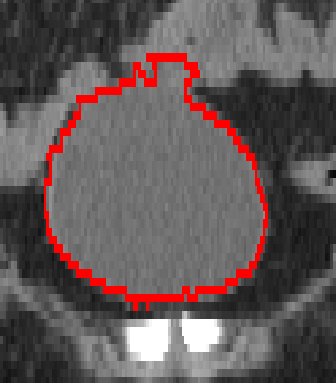} &
		\includegraphics[width=\linewidth]{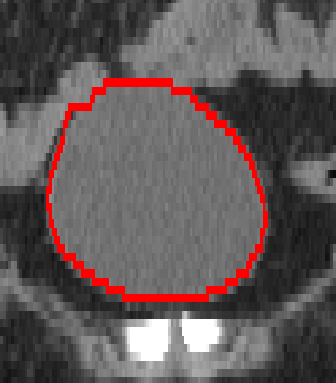} \\
		\rotatebox{90}{axial view} &
		\includegraphics[width=\linewidth]{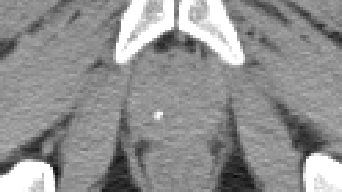} &
		\includegraphics[width=\linewidth]{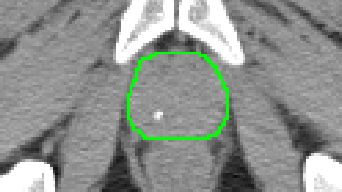} &
		\includegraphics[width=\linewidth]{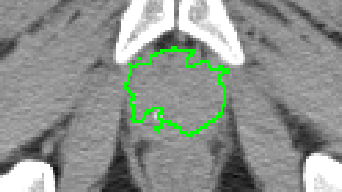} &
		\includegraphics[width=\linewidth]{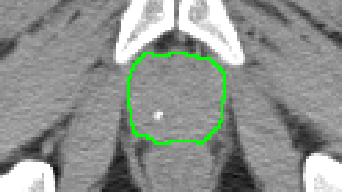} \\
	\end{tabular}
	\caption{An iIllustrative example of segmentations by the proposed method  and the mesh-based LOGISMOS method~\cite{Song2010miccai}. The red and green contours are for bladder and prostate, respectively.}
	\label{fig:prostateContourResult}
\end{figure*}

Fig.\ref{fig:prostateContourResult} shows  example segmentation results of the proposed method and the mesh-based LOGISMOS approach. The LOGISMOS method tends to produce spiky surfaces compared to the smooth appearance of the manual contour. The spiky error at the top of bladder in the mesh-based segmentation (third column in Fig.\ref{fig:prostateContourResult}) signals high risk of mesh folding, which has to be accounted for by special procedures discussed in Sec.\ref{subsec:shapePriorByMesh}. In contrast, the proposed segmentation method with the GVF-based shape priors does not need to handle the mesh self-intersection/folding problem, yielding accurate and smooth segmentation of bladder and prostate. 

\section{\xxx{Discussion}}
\label{sec:discussion}
\xxx{\paragraph{The novelty of the proposed method compared to the star shape and the geodesic star methods} Mathematically, our method shares the same formulation as the star shape method~\cite{Veksler2008} and the geodesic star method~\cite{gulshan_geodesic_2010} to enforce the corresponding shape priors. 
However, our method is more general and flexible, especially for medical image segmentation.}
	
\xxx{The star-shape prior~\cite{Veksler2008} is not flexible enough to handle the complex shapes presented in medical image analysis. The digital rays used to enforce the star-shape prior is not well analogous their corresponding Euclidean rays, with image regions uncovered by the digital rays. This results in no shape control for the corresponding part of the target object. In order to cover a complex shape, a number of star centers need to be put \emph{accurately} inside the object, which can be especially cumbersome for 3D image segmentation. More problematically, the resulting optimization problem (given more than 3 star centers) is computationally intractable~\cite{gibson2014computing}.}
	
\begin{figure}
	\centering
	\includegraphics[width=.5\textwidth]{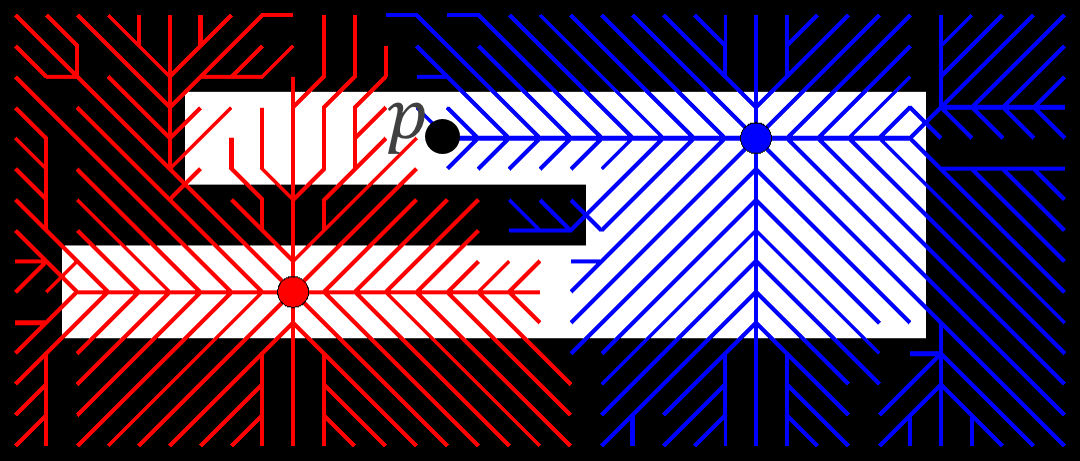}
	\caption{\xxx{Intuitive star center locations of the geodesic star method~\cite{gulshan_geodesic_2010} may lead to undesirable result.}}
	\label{fig:geodesicPath}
\end{figure}	

\xxx{Although the geodesic star method~\cite{gulshan_geodesic_2010} reduces the number of star centers and alleviates the need for accurate star center locations, it still requires a very careful placement of the star centers for segmenting an object with a complex shape. The geodesic star method uses a geodesic distance which combines Euclidean distance with image gradient information. With the user input star centers, the method partitions the image domain into Voronoi cells, each including exactly one star center, such that each voxel in a Voronoi cell is closer to the center of the cell than any other star centers based on its geodesic distances. The assumption here is that the part of the target object in each Voronoi cell is a geodesic star with respect to the star center of the cell. This assumption requires a careful placement of the star centers. In addition, due to the contribution of gradient information in computing geodesic distance, it is sometime counter intuitive to place the star centers for users.
Fig.\ref{fig:geodesicPath} shows an example geodesic forest on a synthetic image, which is generated by putting 0.9 weight on gradient avoidance and 0.1 weight on Euclidean distance, as suggested in~\cite{gulshan_geodesic_2010}. The white region is the target object and the red and blue dots are two star centers intuitively placed by the user. The image is partitioned into two Voronoi cells, the red one and the blue one, based on the geodesic distance. The pixel $p$ is correctly covered by the blue geodesic tree due to the use of the gradient information in its geodesic distance, even though it is closer to the red star center in Euclidean distance. However, the upper left part of the object beyond pixel $p$ is still covered by the red geodesic tree. If we enforce the star shape prior according to that geodesic forest, we would either not be able to get the top left part of the object, or wrongly include the gap in the red Voronoi cell between the upper and lower half of the object as part of the target object.  In summary, if the geodesic star centers are not well-placed, the geodesic star method may fail the segmentation.}
	
	
\xxx{In comparison, our method does not require user to specify seed points (star centers). The pre-segmentation can potentially be obtained by a fully automated method, which saves a large amount of human labor. It is flexible since the pre-segmentation can have complex shapes. The GVF shape prior depends on pre-segmentation shapes instead of raw gradients in original image, and thus is more robust to weak and noisy gradients in medical images. The requirement for pre-segment is also relatively loose: the pre-segmentation can also be either over-segmented or under-segmented or both at different regions. Thus, our proposed GVF shape prior is more suitable to medical image analysis due to the less human labor, flexible shape representation, and robustness to noisy images.}

\xxx{\paragraph{The fidelity to the GVF-based shape prior} 
In our proposed method, the deviation to the shape prior can be penalized by either an infinite or a finite weight, thus achieving the control of the fidelity to the specified shape prior in different degrees.}

\xxx{For our conducted experiments, we set the deviation penalty to an infinite weight due to the flexibility of the GVF-based shape prior. For example, it allows a large range of changes in local surface orientations.  The local surface displacement is also well captured by the GVF shape prior.  With our proposed method, we do scan-wise pre-segmentation to hopefully capture the topology of the target object correctly, and then use the flexible GVF shape prior to recover from moderate amount of pre-segmentation errors. The tolerance of our method to the errors introduced in the shape prior has been demonstrated in Sec.~\ref{sec-brain-sensitivity} by perturbing pre-segmentation. The perturbation to the pre-segmentation basically introduces errors in the shape prior. Fig.~\ref{fig:brainSliceFlexibleGVF}c and Fig.~\ref{fig:brainSliceFlexibleGVF}d visually show how our method can recover from those errors even by using an infinite penalty. }

\xxx{If necessary, we can make use of a finite penalty to enforce the GVF shape prior to allow some GVF paths to trade off the prior information for what the data is telling. However, using a finite penalty introduces one extra parameter to tune, thus one may prefer to use an infinite penalty when the pre-segmentation is not too bad and rely on the flexibility of the GVF shape priors to refine it.  
}

\xxx{\paragraph{Comparison to topology correction algorithm} The topology correction algorithm~\cite{Bazin2007} proposed by Bazin and Pham uses the fast marching technique to propagate the topology of a given template to a binary or probabilistic segmentation. For example, the initial template can be as simple as a 2D donut ring for any single 2D object with one hole in it (the hole in initial template must be aligned with the hole in the object). Their algorithm strictly enforces the classic definition of \emph{global topology}. In contrast, the proposed GVF shape prior enforces strong conformity to \emph{local shape}. For example, the undesired concavity $p_1$ shown in  Fig.\ref{subfig:gvfShapePrior} violates the GVF shape prior. But it would not be corrected by Bazin and Pham's method~\cite{Bazin2007}. Our proposed method shows better local shape resemblance to pre-segmentation besides the global topology. The proposed GVF shape prior enables multiple interactive object segmentation while maintaining the inclusion/exclusion relationship with minimum separation distance. Such interactions are at least nontrivial to be enforced with Bazin and Pham's method.
} 

\section{Conclusion}
We propose a novel GVF-based shape prior which can be directly embedded in the voxel grid space, fundamentally avoiding the mesh folding problem. Its flexibility and power enables to enforce the geometric interactions in multi-object segmentation in an efficient and straightforward fashion.
The proposed segmentation method is validated on the applications of brain tissue segmentation (two nesting/inclusive objects), and of the bladder/prostate segmentation (two exclusive objects). The experiment results demonstrate either superior or competitive segmentation accuracy compared to other state-of-the-art methods. 

\section*{Acknowledgment}
This work was supported in part by the National Science Foundation (NSF) under Grant CCF-1318996, and in part by the the National Institutes of Health (NIH) under Grant R01-EB004640.

\section*{References}

\bibliographystyle{elsarticle-num}
\bibliography{shapePriorBib}







\end{document}